\documentclass[10pt,twocolumn,letterpaper]{article}
 
\usepackage[preprint]{cvpr}

\usepackage{booktabs}
\usepackage{multirow}
\usepackage{bbding}
\usepackage{listings}
\usepackage{xcolor}
\usepackage{mathtools}

\definecolor{cvprblue}{rgb}{0.21,0.49,0.74}
\usepackage[pagebackref,breaklinks,colorlinks,allcolors=cvprblue]{hyperref}

\usepackage{orcidlink}

\newcommand{\mypara}[1]{\noindent\textbf{#1}}

\newcommand{\dataset}{EgoFun3D\xspace}

\definecolor{cbteal}{RGB}{102,194,165}
\definecolor{cborange}{RGB}{252,141,98}
\definecolor{ourpurple}{RGB}{110,39,107}
\definecolor{ourblue}{RGB}{100,175,220}

\newcommand{\receptor}{\ensuremath{{\color{cbteal}\mathcal{R}}}\xspace}
\newcommand{\effector}{\ensuremath{{\color{cborange}\mathcal{E}}}\xspace}
\newcommand{\partfun}{\ensuremath{{\color{red}\mathcal{F}}}\xspace}
\newcommand{\numexp}{\ensuremath{{\color{ourpurple}\mathcal{M}}}\xspace}

\newcommand{\physeff}{\ensuremath{{\color{ourblue}\mathcal{P}}}\xspace}

\definecolor{codegreen}{rgb}{0,0.6,0}
\definecolor{codegray}{rgb}{0.5,0.5,0.5}
\definecolor{codepurple}{rgb}{0.58,0,0.82}
\definecolor{backcolour}{rgb}{0.95,0.95,0.92}

\lstdefinestyle{mystyle}{
  backgroundcolor=\color{backcolour}, commentstyle=\color{codegreen},
  keywordstyle=\color{magenta},
  numberstyle=\tiny\color{codegray},
  stringstyle=\color{codepurple},
  basicstyle=\ttfamily\footnotesize,
  breakatwhitespace=false,         
  breaklines=true,                 
  captionpos=b,                    
  keepspaces=true,                 
  numbers=left,                    
  numbersep=5pt,                  
  showspaces=false,                
  showstringspaces=false,
  showtabs=false,                  
  tabsize=2
}

\usepackage{seqsplit}

\usepackage{chngcntr}

\AtBeginDocument{%
  \renewcommand{\thelstlisting}{\arabic{lstlisting}}%
}

\def\paperID{294} 
\def\confName{3DV\xspace}
\def\confYear{2027\xspace}

\title{EgoFun3D: Modeling Interactive Objects from Egocentric Videos\\using Function Templates}
\author{Weikun Peng \qquad Denys Iliash \qquad Manolis Savva\\
Simon Fraser University\\
\texttt{\color{cyan} \small \href{https://3dlg-hcvc.github.io/EgoFun3D/}{3dlg-hcvc.github.io/EgoFun3D/}}
}
\begin{document}

\newcommand{\figfirstpagefigure}{
\vspace{-3em}
\begin{center}
\captionsetup{type=figure}
\includegraphics[width=0.95\linewidth]{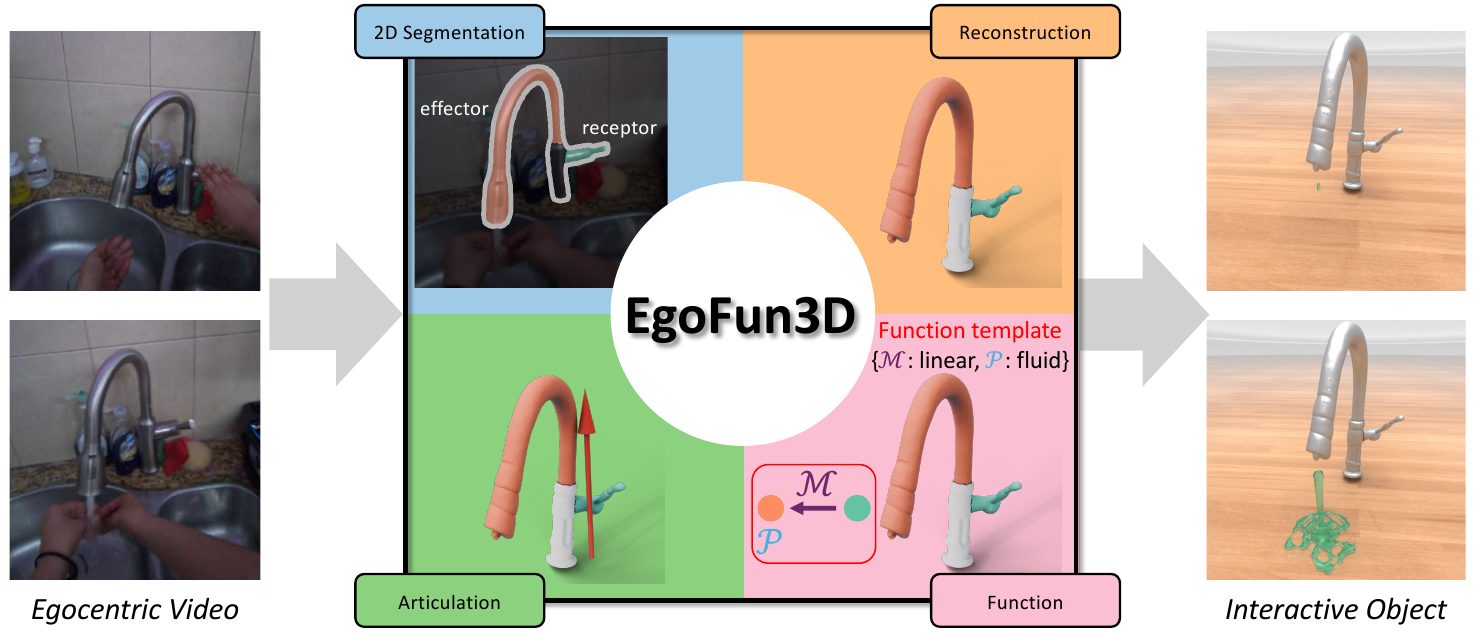}
\captionof{figure}{We present \emph{\dataset}: a coordinated task, dataset and benchmark for modeling interactive 3D objects from egocentric videos. Given an egocentric video as an input, the output is a simulation-ready interactive object (e.g., faucet handle starts water flow from faucet spout). We break down the task into 4 steps to propose a baseline approach using off-the-shelf components. Our \textit{function template} representation, produced by the proposed system, can be compiled into executable code for a simulator of choice.}
\label{fig:teaser}
\end{center}
\vspace{0.5em}
}

\twocolumn[{
    \maketitle
    \figfirstpagefigure
}]
\vspace{-3em}
\begin{abstract}
We present \dataset, a coordinated task formulation, dataset, and benchmark for modeling interactive 3D objects from egocentric videos.
Interactive objects are of high interest for embodied AI but scarce, making modeling from readily available real-world videos valuable.
Our task focuses on obtaining simulation-ready interactive 3D objects from egocentric video input.
While prior work largely focuses on articulations, we capture general cross-part functional mappings (e.g., rotation of stove knob controls stove burner temperature) through function templates, a structured computational representation.
Function templates enable precise evaluation and direct compilation into executable code across simulation platforms.
To enable comprehensive benchmarking, we introduce a dataset of 517 egocentric videos featuring challenging real-world interactions with paired 3D geometry, segmentation over 2D and 3D, articulation and function template annotations.
To tackle the task, we propose a 4-stage pipeline consisting of: 2D part segmentation, reconstruction, articulation estimation, and function template inference.
Comprehensive benchmarking shows that the task is challenging for off-the-shelf methods, highlighting avenues for future work.
\end{abstract}
\vspace*{-14pt}
\section{Introduction}
\label{sec:intro}

Objects composed of functional parts are \textit{interactive}.
Such objects have many applications in gaming, robotics and embodied AI~\cite{jiang2025phystwin, peng2024tiebot, torne2024rialto, dan2025xsim, ning2025prompting, Escontrela25arXiv_GaussGym, abou-chakra2024physically, zhang2025real} but are scarce.
Thus, modeling interactive objects from videos of real-world interactions is of increasing interest.

A special well-studied case of interactive objects is articulated objects, with a number of recent works investigating how to model part motion~\cite{wang2019shape2motion,jiang2022opd,liu2025survey} as well as how to create articulated objects from real-world observations~\cite{peng2025itaco, liu2025videoartgs, jiayi2023paris, weng2024neural, liu2025building}.
However, these works only model movements of parts and not the interaction between parts and physical state changes.
For instance, rotating the stovetop knob changes the temperature of the burner.

While recent attempts model such functionality~\cite{zhang2025open,delitzas2024scenefun3d, hu2026hierarchical}, they typically represent part functionality using natural language which is neither formal nor precise enough to enable structured evaluation or reliable conversion to simulation-ready assets.
Other works~\cite{kolve2017ai2} specify action state change using simulator APIs.
While these simulators allow implementing executable part functions~\cite{Genesis,NVIDIA_Isaac_Sim,li2023behavior}, the APIs and code are heterogeneous and do not provide a unified and portable representation of functionality.

\begin{figure}[t]
    \centering
    \includegraphics[width=0.95\linewidth]{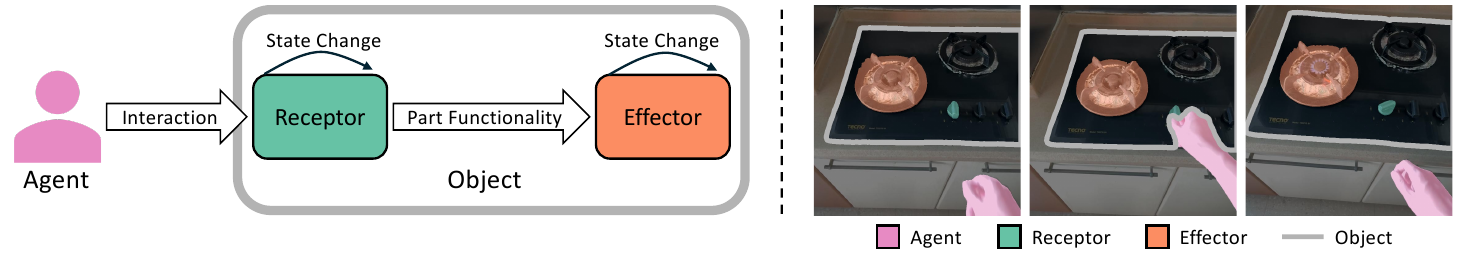}
    \caption{Illustration of a typical human-object interaction. An agent interacts with a receptor, changing its state. \emph{Part functionality} defines how the state change of the receptor maps to the state change of the effector. The example on the right shows a human interacting with a stovetop knob. The part function triggers the temperature change of the burner after knob actuation.}
    \vspace{-12pt}
    \label{fig:formulation}
\end{figure}

To address this, we introduce a general notion of \textit{part functionality} that maps how actions on one part (the \emph{receptor}) lead to a response in another part (the \emph{effector}).\footnote{We take this terminology from biological reflex arcs~\cite{sherrington2023integrative} where receptors detect a stimulus and effectors produce an action in response.}
For example, pulling the faucet handle opens the faucet spout and achieves the response of water flowing (\cref{fig:teaser}).
We propose \emph{function templates}, a unified representation of part functionality that enables evaluation and compilation into code.
The function template abstracts an object into the receptor part that is manipulated and the effector part that produces a response (see \cref{fig:formulation}).
Function templates have two key components: \emph{mapping} and \emph{physical effect}.
The mapping specifies how the receptor state influences the effector state.
The physical effect specifies the physical phenomenon and units of the effector response.

Armed with our function template abstraction, we study how to build interactive objects that can be used in simulation.
We formulate and tackle the new task of modeling interactive objects using function templates from egocentric video data,
which are increasingly abundant and capture human-object activities with common interactive objects~\cite{grauman2024ego, perrett2025hdepic}.
This task poses many challenges, as egocentric videos feature \emph{highly dynamic viewpoints, severe occlusions, and complex interactions}.
As no prior dataset contains all the modalities required to benchmark this task (see \cref{tab:dataset comparison}), we create a benchmark dataset featuring: 1) real-world, egocentric videos; 2) 3D meshes of objects; 3) object and part segmentation annotations on both 2D and 3D; 4) part articulation annotations; and 5) function template annotations.

\begin{table*}[t]
    \vspace{-12pt}
    \caption{Comparison of egocentric human-object interaction video datasets.
    \dataset provides real egocentric videos and complete geometry with the richest annotation compared to datasets from prior work. --- indicates annotations not present due to missing modalities.
    }
    \vspace{-6pt}
    \centering
    \setlength{\tabcolsep}{8pt}
    \resizebox{\textwidth}{!}{
    \begin{tabular}{@{} l c c c c c c c c c @{}}
    \toprule
        \multirow{2}{*}{Dataset} & \multirow{2}{*}{\shortstack{Object\\3D geometry}} & \multirow{2}{*}{\shortstack{Physical\\effect types}}  & \# Ego. Videos & \# Objects & \# Obj. Categories & \multicolumn{4}{c}{Annotations} \\
         \cmidrule(l{2pt}r{2pt}){7-10}
          & & & & & & 2D Part Masks & 3D Part Masks & Articulation & Part Functionality \\
         \midrule
         Ego-Exo4D~\cite{grauman2024ego} & --- & --- & 5035 & --- & --- & \textcolor{red}{\XSolidBrush} & --- & --- & \textcolor{red}{\XSolidBrush} \\
         HD-Epic~\cite{perrett2025hdepic} & cuboid primitive & --- & 156 & 308 & 21 & \textcolor{red}{\XSolidBrush} & \textcolor{red}{\XSolidBrush} & \textcolor{red}{\XSolidBrush} & \textcolor{red}{\XSolidBrush} \\
         Hoi!~\cite{engelbracht2025hoi} & scanned point cloud & articulation (1) & 26 & 112 & 10 & \textcolor{red}{\XSolidBrush} & \textcolor{green}{\CheckmarkBold} & \textcolor{green}{\CheckmarkBold} & \textcolor{red}{\XSolidBrush} \\
         Arti4D~\cite{arti25werby} & scanned point cloud & articulation (1) & 45 & 63 & 14 & \textcolor{red}{\XSolidBrush} & \textcolor{red}{\XSolidBrush} & \textcolor{green}{\CheckmarkBold} & \textcolor{red}{\XSolidBrush} \\
         \midrule
         \textbf{\dataset} (ours) & real mesh with interior & various (5) & 517 & 128 & 32 & \textcolor{green}{\CheckmarkBold} & \textcolor{green}{\CheckmarkBold} & \textcolor{green}{\CheckmarkBold} & \textcolor{green}{\CheckmarkBold} \\
         \bottomrule
    \end{tabular}}
    \vspace{-12pt}
    \label{tab:dataset comparison}
\end{table*}

We propose a baseline approach for the task, powered by off-the-shelf components in a 4-step breakdown: 1) segmenting the receptor and effector; 2) reconstructing the geometry of the receptor and effector; 3) inferring how the parts articulate; 4) inferring a suitable function template.
We then use our dataset for extensive benchmarking.
We find that objects with small, severely occluded parts in highly dynamic videos pose a challenge for every established off-the-shelf component in our baseline approach.

In summary, our contributions include:
1) Formalizing the task of modeling interactive 3D objects from egocentric videos;
2) Introducing the function template representation to capture general part functionality and obtain simulation-ready objects;
3) Creating a dataset for comprehensive end-to-end benchmarking of off-the-shelf components on this task;
4) Proposing a baseline approach for the task using the best components identified by our benchmarking.
\section{Related Work}
\label{sec:related works}

\mypara{Articulated object modeling.}
Articulated objects are a common type of interactive object.
Modeling articulated objects has been studied for years.
Most works focus on reconstructing interactive rigid or articulated objects from static scans~\cite{jiayi2023paris, weng2024neural, liu2025building, chen2024urdformer, mandi2024real2code, jiayi2024singapo, li2025art, yuan2025larm, chen2025artilatent, chen2025freeart3d}.
Recent works have begun using videos of manipulating articulated objects as input to build interactive articulated objects in simulation~\cite{le2024articulate, peng2025itaco, arti25werby, liu2025videoartgs, delitzas2026funrec, guo2026articulat3d}.
These approaches take videos as input and output part-level reconstruction and articulation parameters.
However, their focus is only on articulated parts, ignoring other functions such as illumination.
Thus, these methods cannot be applied directly to our problem.

\mypara{Part functionality modeling.}
Recent work has studied more general functional relationships between parts. SceneFun3D~\cite{delitzas2024scenefun3d} introduces functionality segmentation and task-driven affordance grounding.
One line of follow-up work studies how to better understand part functionality and affordance~\cite{corsetti2025functionality, liu2024grounding} by
outputting affordance segmentations on the point cloud.
Another line of work attempts to build a functional 3D scene graph to represent the functional relationship between parts and objects~\cite{zhang2025open, rotondi2025fungraph, gu2025artisg, hu2026hierarchical}.
MoMa-SG combines building 3D scene graphs and articulation estimation together to assist mobile manipulation~\cite{buechner2026momasg}.
This line of work is closer to our setting, but it takes a 3D scene understanding perspective, whereas we model interactive 3D objects.
Moreover, they use natural language whereas we extract function templates that enable portable instantiation as executable functions across simulators.

\mypara{Datasets for part functionality.}
At the scene level, SceneFun3D~\cite{delitzas2024scenefun3d} annotates interactive elements in 3D scenes with masks and natural language descriptions of their functions.
FunGraph3D~\cite{rotondi2025fungraph} provides scene graphs to describe the functional relationship between different object parts. HHOpenFunGraph~\cite{hu2026hierarchical} extends FunGraph3D with hierarchical functional relationship.
Articulate3D~\cite{halacheva2024articulate3d} provides articulation parameters and part segmentation in 3D scenes.
These works focus on evaluating functionality understanding from a point cloud or a video of the static scene.
At the object level, prior work have introduced synthetic datasets that  describe part functionality in natural language~\cite{cao2025physx} or implement executable part functionality in simulation~\cite{li2023behavior, jin2025artvip}.
However, they do not target evaluation of modeling interactive objects from real videos. 
Arti4D~\cite{arti25werby} and iTACO~\cite{peng2025itaco} evaluate estimating articulated parameters from RGBD videos.
Hoi!~\cite{engelbracht2025hoi} collects videos of manipulating articulated objects accompanied by force annotations and scanned point clouds of the objects.
These datasets are the closest to our work, but they consider only rigid or articulated objects, whereas we model object interactions with a broader set of physical state changes.
\Cref{tab:dataset comparison} compares the most relevant datasets and their properties.
\section{Methodology}
\label{sec:problem formulation}

\subsection{Function Templates}
\label{subsec:task_representation}

Function templates define part functionality based on coupling a receptor and effector part (see \cref{fig:formulation}):
1) An agent actuates the receptor;
2) The receptor state change is mapped to the state change of the effector;
3) The effector state change causes further state changes in the environment.
The goal of the agent is not to actuate the receptor, but to change the effector through actuating the receptor.

We denote the states of the receptor \receptor and effector \effector by $s_{\effector}$ and $s_{\receptor}$ respectively.
Then, the relation between receptor and effector is defined as $s_{\effector} = \partfun(s_{\receptor})$. Function templates decompose the \partfun function into physical effect \physeff and response mapping \numexp.
Since the specifics of mapping (i.e., conversion of units) depend on physical effect, we instantiate the function template as $\partfun(\cdot) \coloneq \numexp_{\physeff}(\cdot)$.

In egocentric videos of indoor scene interaction scenarios, a few physical effects dominate. 
These physical effects are changes of:
1) \textbf{geometry} (e.g. door opening);
2) \textbf{illumination} (e.g. light emission);
3) \textbf{temperature} (e.g. stove burner heating up);
4) \textbf{fluid} (e.g. faucet spout releasing water) and
5) \textbf{air flow} (e.g. hair dryer blowing air).
While conceptually \physeff can represent any physical effect, we target modeling these five as a foundation, as they are widely-present, supported by simulation platforms, and result in visually-observable changes. We observe that these five physical effects already cover 88.2\% of all functional relationships in the FunGraph3D dataset~\cite{zhang2025open}.

\begin{figure}[t]
    \centering
    \includegraphics[width=0.95\linewidth]{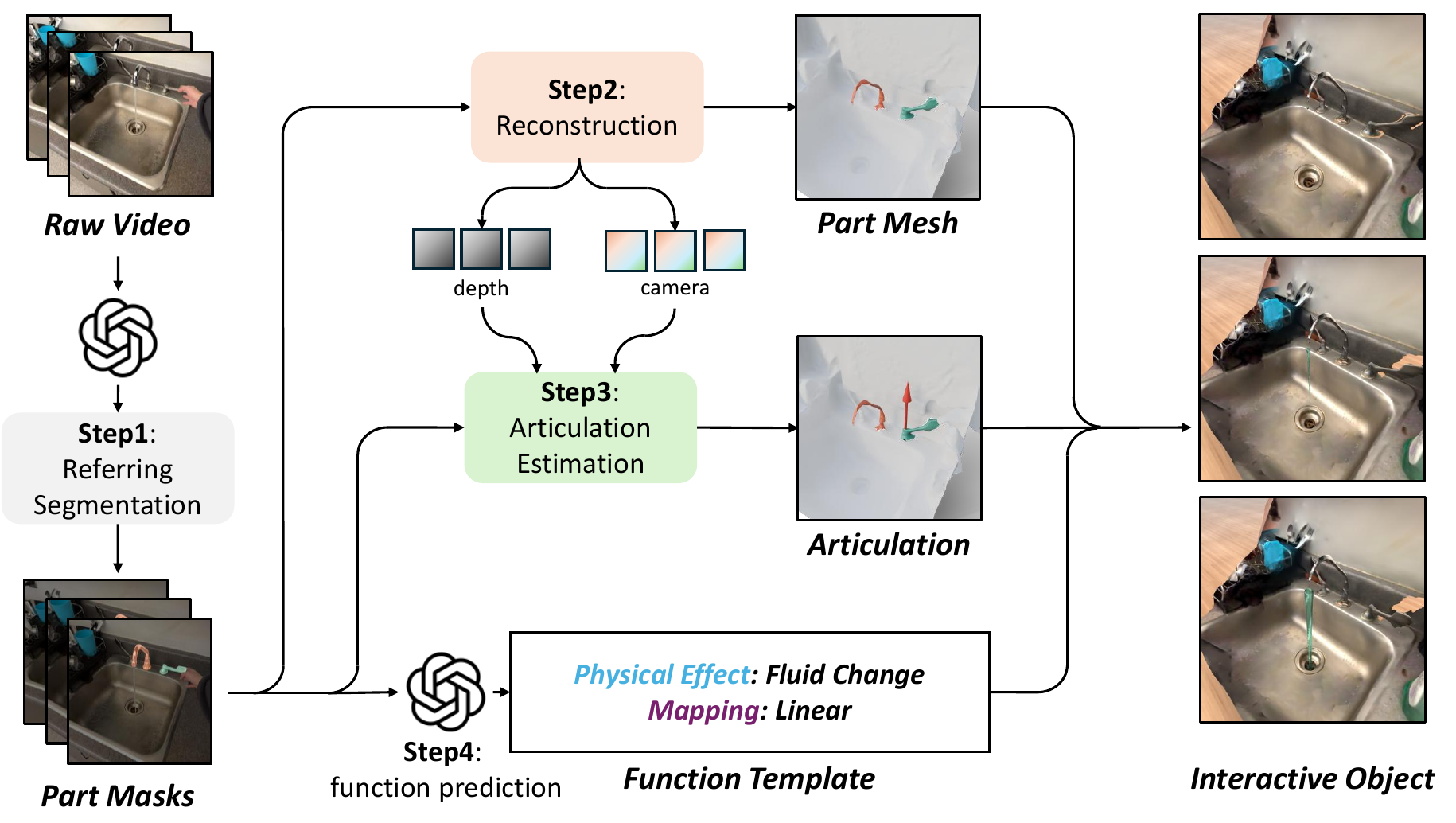}
    \caption{Our baseline framework. We break down the task into 4 steps that are individually targeted with off-the shelf components. First, a VLM generates part descriptions which are used to segment the parts in the video. Then, the geometry of the receptor and the effector are reconstructed, articulation parameters are estimated, and the function template is inferred. These outputs are combined to build the interactive object.}
    \vspace{-12pt}
    \label{fig:method}
\end{figure}

\begin{figure*}[t]
    \centering
    \vspace{-10pt}
    \includegraphics[width=0.9\linewidth]{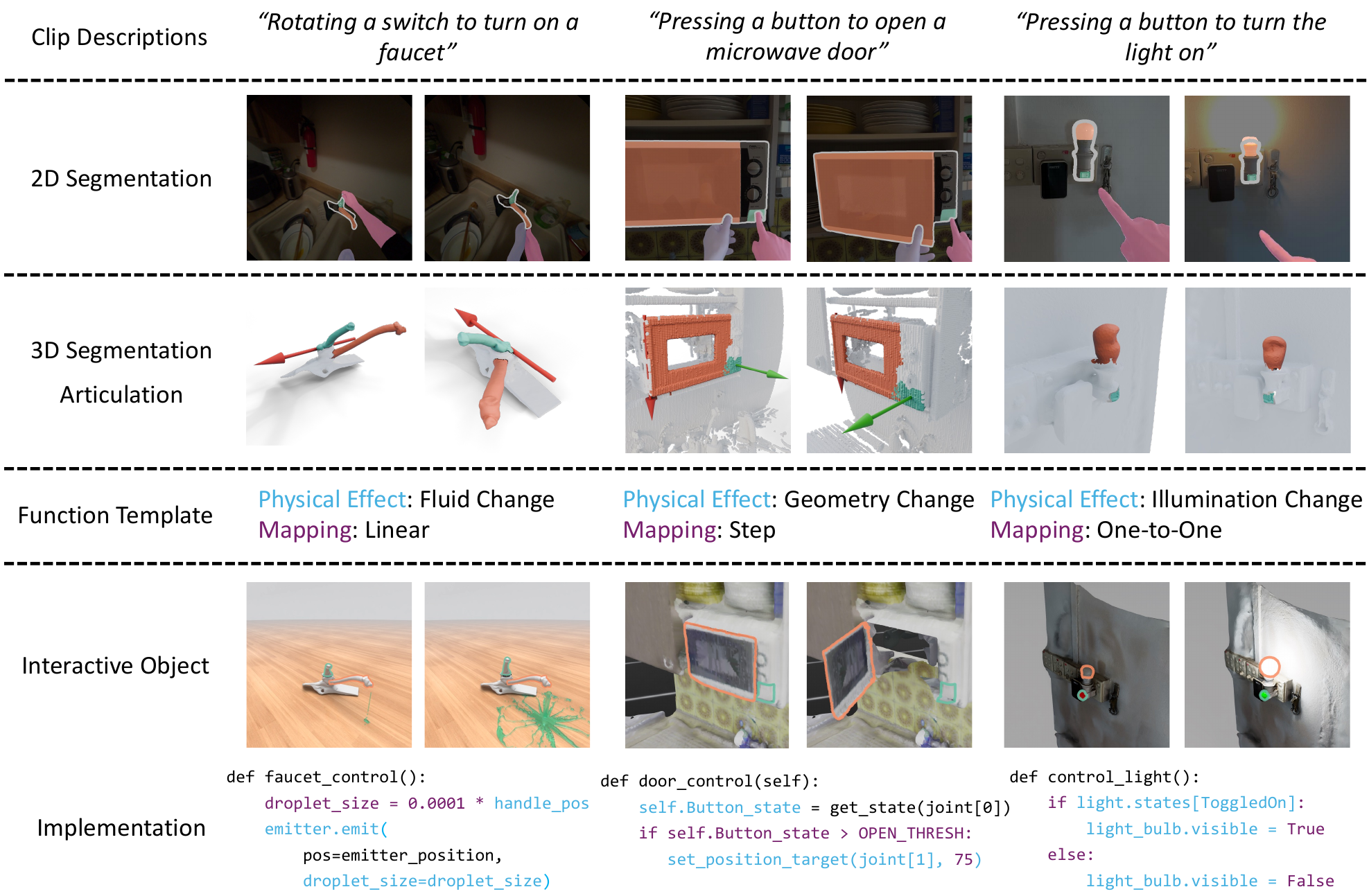}
    \vspace{-10pt}
    \caption{Examples of our annotations. We provide 2D segmentation masks for hands, receptor (\textcolor{cbteal}{teal}), effector (\textcolor{cborange}{orange}), and the complete object. We annotate part segmentation for receptor and effector on 3D meshes. We also annotate revolute and prismatic joints, shown as \textcolor{red}{red} and \textcolor{Green}{green} arrows respectively. For the function template, we specify a physical effects and a mapping. Finally, we create concrete instantiations of interactive objects in different simulators: Genesis\cite{Genesis} (left), Isaac Sim~\cite{NVIDIA_Isaac_Sim} (middle), BEHAVIOR~\cite{li2023behavior} (right).}
    \vspace{-8pt}
    \label{fig:annotation}
\end{figure*}

Annotating mappings for each receptor and effector pair is challenging.
First, accurate measurement of physical parameters such as illumination intensity, stove temperature, and water flow speed is impractical.
Second, relevant parameters in simulation platforms do not directly correspond to physical measurements and vary across simulators.
For example, water flow speed for a faucet can be measured as the volume of water per second but it is typically parameterized by the number and speed of fluid particles~\cite{Genesis}.
Due to these challenges, we abstract the mapping into a few types that strike a balance between generality and concreteness.

In practice, we observe four common types of mappings in our data which we represent using appropriate mathematical functions:
1) \textbf{binary} (e.g., button-type light switch turning on room light);
2) \textbf{step} (e.g., fridge door turning off fridge light at specific angle)
3) \textbf{linear} (e.g., knob-type light switch controlling lamp light intensity)
4) \textbf{cumulative} (e.g., electric stove with buttons controlling the increase or decrease of the burner's temperature).
See the supplement for a more detailed formalization.

The above formalization distills arbitrary real-world \partfun into two specific components: the physical effect \physeff and a response mapping \numexp that can be modeled computationally.
The resulting function templates provide a unified representation that enables portability across simulators and easy evaluation. Moreover, due to the compact design of function templates, it is easy to add new function templates or compose existing function templates to model more complex functionality.
Arbitrary combinations of components in the function templates can be compiled into executable Python scripts for specific simulation platforms. For example, binary and step functions can be represented using an if-else block, while linear and cumulative functions using simple linear equations with relevant physical parameters according to the physical effect $\physeff$.
In this paper, we show example outputs targeted to BEHAVIOR~\cite{li2023behavior}, Isaac Sim~\cite{NVIDIA_Isaac_Sim}, and Genesis~\cite{Genesis} (see \cref{fig:annotation}).

\subsection{Task Definition}
\label{subsec:problem_setting}

After formalizing function templates, we can now define the task.
The input egocentric video $\mathcal{V}$ captures an agent interacting with an object.
The output is an interactive object, parametrized by $\{ (\mathcal{G_{\receptor}}, \mathcal{A}_{\receptor}),  (\mathcal{G_{\effector}}, \mathcal{A}_{\effector}), \numexp_{\physeff} \}$, where $\mathcal{G}$ are per-part 3D meshes. $\mathcal{A}_{\cdot}$ is an articulation parametrized by a tuple $(J_t, J_a, J_o, J_r)$, where $J_t \in \{\mathtt{fixed, prismatic, revolute}\}$ is a joint type, $J_a \in \mathcal{R}^3$ is the joint axis, $J_o \in \mathcal{R}^3$ is the joint origin and $J_r \in \mathcal{R}^{2 \times 1}$ are the minimum and maximum values of range of motion. $\numexp_{\physeff}$ is a function template as defined previously in \cref{subsec:task_representation}, with $\numexp \in \{\mathtt{binary, step, linear, cumulative}\}$ and {\footnotesize $\physeff \in \{\mathtt{\seqsplit{geometry, illumination, temperature, fluid, airflow}}\}$}.
\subsection{Approach Framework}
\label{sec:method}

We propose to break the task down into 4 steps that can be implemented by off-the-shelf components: 2D segmentation, reconstruction, articulation estimation and function template inference. The proposed system is visualized in \cref{fig:method}. First, we identify which parts are involved in the interaction and their functional roles. We prompt a VLM to identify the receptor and effector and describe them in detail. We then use these descriptions with a 2D referring segmentation model to obtain the part masks. We reconstruct the geometry of the receptor and effector by leveraging a reconstruction model combined with the predicted part masks. Such models also estimate the depth maps, camera intrinsics and extrinsics, which are used next by the articulation estimation model. We use a VLM once again to infer the mapping and physical effect, populating the function template. Finally, function templates are compiled into executable code and, along with part geometry and articulations, result in a simulation-ready interactive object.

\section{EgoFun3D Dataset}
\label{sec:dataset}

No prior dataset is suitable for evaluating our task end-to-end.
Therefore, we create a dataset to enable benchmarking.
Our EgoFun3D dataset is built from egocentric videos segmented into short interaction snippets.
For each snippet, we provide a natural language description of the action, and 2D segmentation masks of hands, the object interacted with, and receptor and effector over all video frames.
We also provide 3D receptor and effector masks over the 3D object meshes, 3D articulation parameters, and function templates relating the action on the effector to the receptor (see \cref{fig:annotation}).
We describe how the data is curated (\cref{subsec:data_curation}), annotated (\cref{subsec:data_annotation}), and give summary statistics (\cref{subsec:data_stats}).

\subsection{Data Curation}
\label{subsec:data_curation}
  
Building a dataset for our task requires annotations across egocentric videos and 3D object meshes.
Since the combined modalities are not available in any prior dataset, we collect a set of egocentric video as the basis for our dataset.  
\mypara{Videos.} We curate a set of videos from Ego-Exo4D~\cite{grauman2024ego} and FunGraph3D~\cite{zhang2025open}.
For improved coverage of object categories with different physical effects, we capture new raw egocentric videos mostly with Aria Gen 1 glasses~\cite{engel2023project} and some with iPhone 11 Pro using Record3D~\cite{record3d}. 
For our self-captured videos, we also collect posed images of the interacted object using Polycam~\cite{polycam}.
We manually split each video into short clips containing a single interaction, and describe the observed action in natural language (e.g., ``rotating a switch to turn on the faucet'').

\mypara{Object geometry.} We obtain the 3D geometry using a workflow leveraging multiple approaches with a priority ordering, and manual quality verification at each step.
We first attempt to find CAD model with permissive licensing from the official product website of each object.
If that fails, the next priority is reconstruction from object-centric posed images using MP-SfM~\cite{pataki2025mpsfm}, GeoSVR~\cite{li2025geosvr} as well as a proprietary Polycam~\cite{polycam} pipeline.
If the results of these methods are inaccurate (common for reflective or transparent surfaces etc.), we proceed to use generative models to obtain the geometry.
We use MV-SAM3D~\cite{li2026mv}, TRELLIS.2~\cite{xiang2025trellis2} and ShapeR~\cite{siddiqui2026shaperrobustconditional3d} to produce the candidates.
In particular, ShapeR is used for the videos sourced from Ego-Exo4D as object-centric posed images are not available, but a sparse point cloud which can be used with ShapeR is available.
If ShapeR fails to generate an accurate mesh, we select a timestamp in the video and use both egocentric and exocentric views to build the object mesh using MapAnything~\cite{keetha2025mapanything}.
As FunGraph3D provides high quality point clouds, we run mesh reconstruction on them directly using \cite{Liu2024DWG}.

The best output of the workflow is selected and manually refined in Blender~\cite{blender} to remove redundant geometry, add missing surfaces, and merge parts from different methods, etc.
See the supplement for more details.

\subsection{Data Annotation}
\label{subsec:data_annotation}

Our data annotation addresses 4 aspects: 2D segmentation masks for hands, object, and parts; 3D part segmentation masks; articulation parameters; and function templates.

\mypara{2D segmentation:} we manually add point prompts for objects and their parts in about 10 frames per video and propagate them to all frames using SAM2~\cite{ravi2024sam2}. We manually check the propagated masks and iterate to fix the issues. We annotate the segmentation for left and right hands, receptor and effector, as well as segmentation of the entire object.
\mypara{3D segmentation:} we oversegment the mesh using Felzenszwalb and Huttenlocher's graph-based algorithm~\cite{felzenszwalb2004efficient} and assign them to receptors and effectors.

\mypara{Articulation parameters:} we manually annotate joint types, axes, ranges and origins (for revolute joints).

\mypara{Function templates:} are built manually by selecting appropriate mapping and physical effect pairs. Episodes exhibiting complex functionality (i.e., configuring a washing machine program) are labeled as ``miscellaneous''.
\cref{fig:annotation} shows example data samples with annotations. 

\begin{figure}[t]
    \centering
    \includegraphics[width=0.95\linewidth]{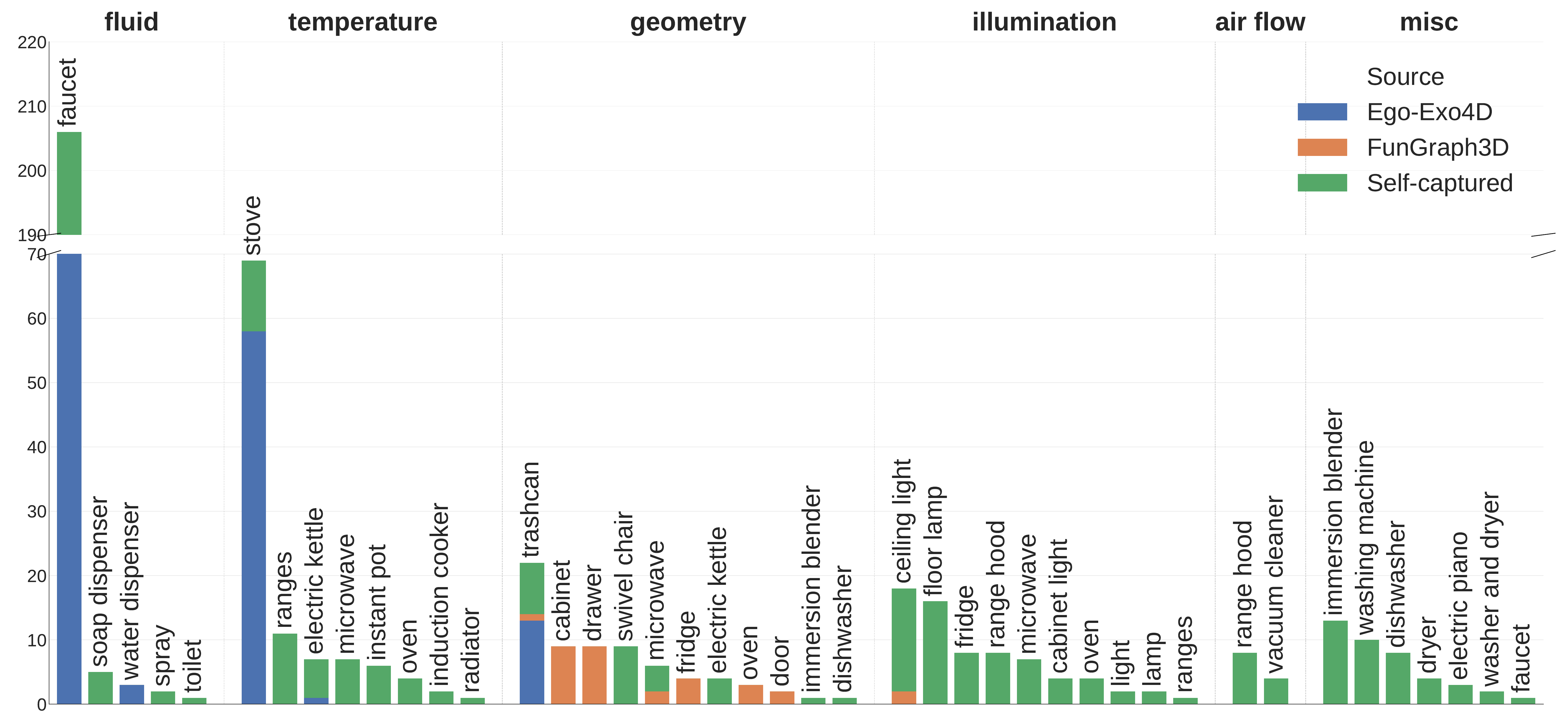}
    \includegraphics[width=0.95\linewidth]{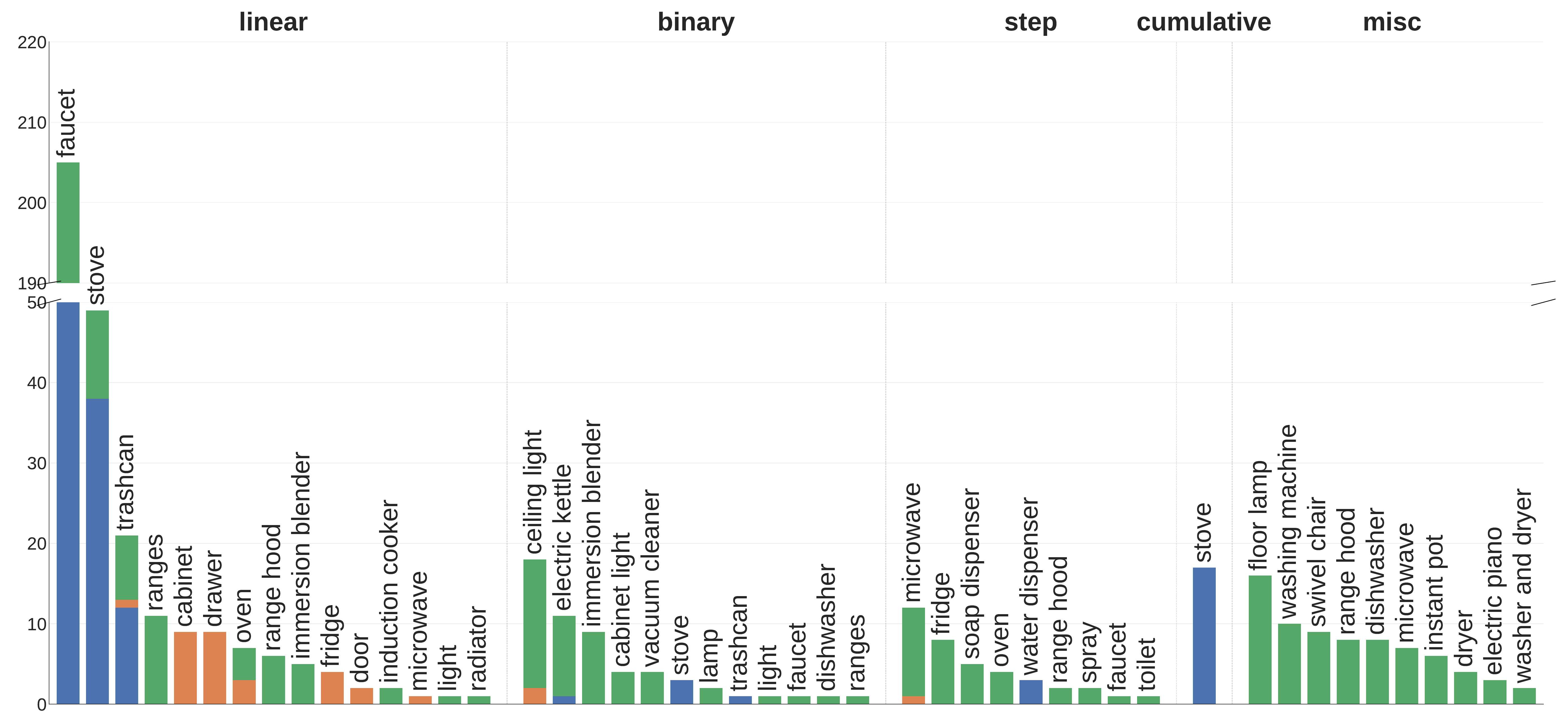}
    \caption{Number of egocentric videos per object category, grouped by physical effect (top) and by function mapping (bottom). Although there are prominent long-tail distributions across categories, physical effects, and function mappings, self-captured data provides a more balanced distribution and diverse object categories, as shown in green.}
    \vspace{-12pt}
    \label{fig:video distribution}
\end{figure}

\subsection{Dataset Statistics}
\label{subsec:data_stats}

In total, we collect 517 egocentric videos 
(259 self-captured videos, 226 from Ego-Exo4D, and 32 from FunGraph3D), 
featuring interactions with 128 different object instances across 32 different categories.
We captured videos to better cover diverse object categories and physical effects (Ego-Exo4D and FunGraph3D videos only cover 5 and 8 object categories respectively). 
\cref{fig:video distribution} shows the data distributions.
Data samples featuring faucets and stoves make up the majority of our dataset, as much of our data are captured in kitchen scenarios.
This scenario is among the most functionally-complex common human activities except for interactions with computing devices.
The data bias can also lead to a skewed distribution for functional mappings and physical effects as faucets and stoves are naturally biased towards certain combinations.
The videos we source from existing datasets are particularly prone to this bias, whereas the videos we capture have a more balanced distribution.

\section{Experiments}
\label{sec:experiments}

We benchmark off-the-shelf components that implement the 4 steps of our task breakdown. In this section, we discuss the methods we benchmark for each step (\cref{subsec:exp_baselines}), the evaluation protocol (\cref{subsec:exp_metrics}) and our findings (\cref{subsec:exp_results}).

\subsection{Baselines and Implementation Details}
\label{subsec:exp_baselines}

\mypara{2D segmentation.}
We first use Gemini 3 Flash~\cite{geminiteam2025gemini3} to output a short label and a long description of the receptor and effector in the video clip. The long description provides spatial relationships (e.g., the \textit{right-most} knob) to disambiguate the instance with which the human interacts in the video, as multiple parts with the same label might be present. See the supplement for the VLM prompting details.
We then select a few representative methods for referring expression segmentation grouped into 2 categories: those natively supporting text prompts, and those requiring geometric prompts (i.e., clicks, bounding boxes).
X-SAM~\cite{wang2025xsam}, Sa2VA~\cite{yuan2025sa2va} and SAM3~\cite{carion2025sam3segmentconcepts} are representative approaches that support text prompts natively.
SAM3, however, is not trained to take long descriptions as input.
Hence, we pair it with Qwen3-VL 8B~\cite{Qwen3-VL} following the SAM3-Agent setup in the public SAM3 codebase. This setup establishes an agentic framework where the VLM uses SAM3 as a tool.
We also include a few models that output geometric prompts based on natural language descriptions, such as Molmo2~\cite{clark2026molmo2} and VisionReasoner~\cite{liu2026visionreasoner} and combine them with SAM for segmentation.
To achieve feasible runtimes, we subsample 20 frames from each video uniformly and segment them.
Then, we use SAM3 to propagate the segmentation to all video frames, with the initial 20 acting as prompts.
See the supplement for more details.

\mypara{Reconstuction.}
We select three representative methods for reconstruction: MapAnything~\cite{keetha2025mapanything}, Depth Anything 3~\cite{depthanything3}, and ViPE~\cite{huang2025vipe}.
MapAnything and Depth Anything 3 are multi-view transformers that reconstruct 3D scenes from a series of images.
ViPE is a more complex, multi-stage optimization framework.
For MapAnything and Depth Anything 3, we divide the full video into several chunks with 20 frames for each chunk. Then, we reconstruct each chunk of the video individually and stitch the results of all chunks for computational feasibility.
ViPE handles frame downsampling, so we provide it the full video directly.
The outputs require a post-processing step to fuse the point clouds into a single state, as the input videos feature moving parts.
We use RoMa~\cite{edstedt2024roma} for pair-wise feature matching to estimate rigid transformations between the frames and fuse the point cloud to the state of the first frame.
Finally, we aggregate reconstructed meshes for each part across all frames to obtain the final reconstruction.
See the supplement for details.

\mypara{Articulation estimation.}
ArtiPoint~\cite{arti25werby} and iTACO~\cite{peng2025itaco} are two recent methods for articulation estimation.
Artipoint uses point tracks to build a factor graph and estimate articulation parameters.
We remove hand detection and object segmentation modules, as segmentation masks in our setting are provided from an upstream prediction.
iTACO first estimates the articulation parameters using image feature matching and then refines the initial proposal using gradient-based optimization.
We remove camera pose and moving part estimation modules from iTACO as those are provided from the upstream modules.

\mypara{Function template inference.}
We compare four VLMs for function template estimation: Gemini 3 Flash~\cite{geminiteam2025gemini3}, GPT-5 mini~\cite{openai2025gpt5}, Molmo2 8B~\cite{clark2026molmo2}, and Qwen3-VL 8B~\cite{Qwen3-VL}.
A video with receptor and effector parts highlighted in different colors is the input and we prompt VLM to infer 1 of 5 physical effects and 1 of 4 mappings.

\subsection{Evaluation Protocol}
\label{subsec:exp_metrics}

\textbf{Segmentation.} We use IoU to evaluate the segmentation. The results are reported on masks propagated to all frames. As downstream pipeline steps rely on segmentation, we select only the parts with average IoU greater than 50\% for evaluation of the downstream modules. We consider such cases to be a segmentation success, and report the success rates. The evaluation of downstream modules takes the successful masks predicted by the best method as input. This is done to decouple the effects of severely incorrect segmentation results from the raw performance of downstream modules while evaluating the performance of the entire pipeline considering reasonable error accumulation. 

\mypara{Reconstruction.} We use the chamfer distance to evaluate reconstruction of the receptor and effector. We report the median of the distance distribution of each part type separately, as well as across both part types combined. The choice of a median, in contrast to the mean, is due to occasionally-produced severely corrupted predictions that are capable of skewing the mean. See supplement for more discussion. We also report the mean camera pose error to assist analysis.

\mypara{Articulation.} We use depth maps, camera intrinsics and extrinsics predicted by the best reconstruction method we identify, as input to articulation estimation models. Articulation estimation is evaluated with the joint type accuracy, joint axis error in radians, joint origin error in meters (for revolute joints only). The failure rate count when methods have no output.

\mypara{Function templates.}
For function templates, we report accuracies for classifying physical effect and mapping, as well as an accuracy of predicting both at the same time. Function templates labeled as ``miscellaneous'' are not included in the evaluation of function template prediction.

\mypara{Experimental details.}
We use Nvidia L40S GPUs to run our experiments and report runtimes.
See supplement for evaluation of each module in oracle settings.

\subsection{Experimental Results}
\label{subsec:exp_results}

\begin{figure}[t]
    \centering
    \includegraphics[width=0.95\linewidth]{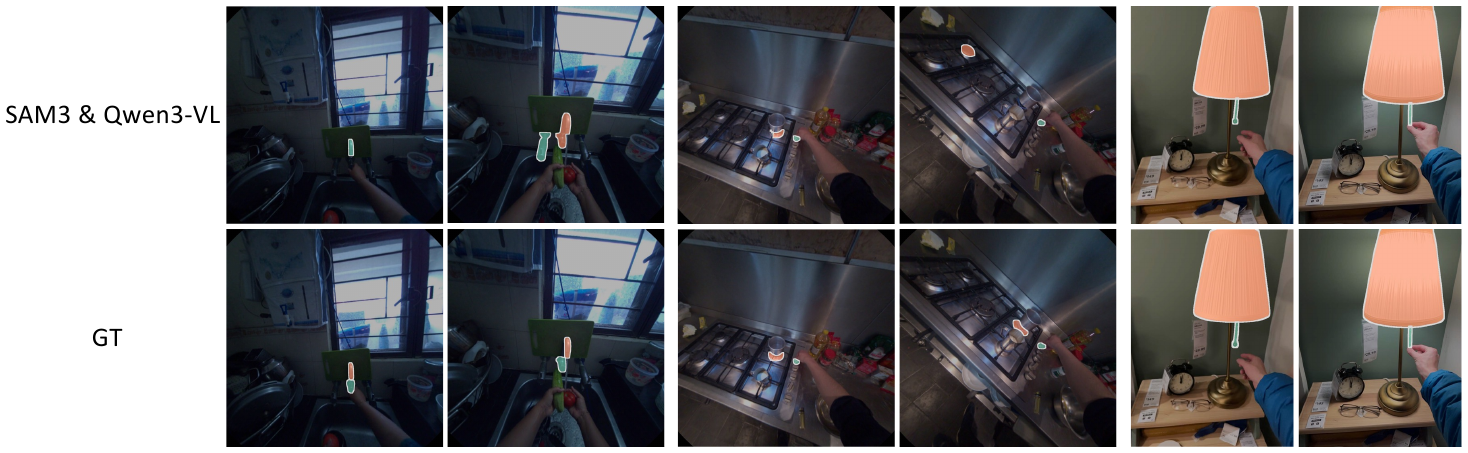}
    \caption{Example 2D segmentation results. We find that SAM3 with Qwen3-VL provides the best segmentation. The main challenges in this subtask are segmentation of incorrect parts (left) and confusion between part instances across frames (middle). Performance on videos featuring more static viewpoints and no part instance ambiguity is better, though such videos are rare (right).}
    \vspace{-12pt}
    \label{fig:segmentation qualitative}
\end{figure}

\mypara{Segmentation consistency and small parts are bottlenecks.}
Overall, the segmentation performance is unsatisfactory (\cref{tab:segmentation}).
We find that using SAM3 with Qwen3-VL~\cite{Qwen3-VL} is the best option by a large margin.
Hence, its outputs are used as inputs when evaluating the downstream modules.
VisionReasoner strikes a better balance between performance and runtime.
In general, segmenting receptors is much more challenging than segmenting effectors, as receptors tend to be smaller parts such as stove knobs and faucet handles.
\cref{fig:segmentation qualitative} shows qualitative examples and the supplement provides more quantitative comparisons.
We see that common failure modes include segmentation of incorrect parts and inconsistent segmentation of different instances across frames.
The results suggest that the agentic framework for 2D segmentation with extensive reasoning is more promising, particularly for challenging scenarios.

\begin{table}[t]
    \caption{Evaluation of 2D segmentation performance. We find that SAM3 \& Qwen3-VL outperforms other methods by a large margin, but is very inefficient.}
    \centering
    \small
    \resizebox{\linewidth}{!}{
    \begin{tabular}{@{} l c c c c c r @{}}
        \toprule
        \multirow{2}{*}{\textbf{Model}}
        & \multicolumn{3}{c}{\textbf{IoU (\%)}}
        & \multicolumn{2}{c}{\textbf{Success (\%)}}
        & \multirow{2}{*}{\shortstack{\textbf{Avg.}\\\textbf{Run. (s)}}} \\

        \cmidrule(lr){2-4} \cmidrule(lr){5-6}

        & Receptor & Effector & Avg.
        & Receptor & Effector
        & \\

        \midrule

        VisionReasoner & 15.4 & 28.3 & 21.8 & 8.4 & 25.8 & 127  \\
        SAM3 \& Qwen3-VL & \textbf{28.2} & \textbf{38.9} & \textbf{33.6} & \textbf{22.2} & \textbf{43.9} & 1769 \\
        SAM3 \& Molmo2 & 16.7 & 27.5 & 22.1 & 8.2 & 22.4 & 92 \\
        X-SAM & 4.0 & 13.7 & 8.8 & 1.5 & 8.4 & \textbf{22} \\
        Sa2VA & 15.8 & 33.6 & 24.7 & 6.0 & 27.7 & 865 \\

        \bottomrule
    \end{tabular}
    }
 \label{tab:segmentation}    
\end{table}

\begin{figure}[t]
    \centering
    \includegraphics[width=0.9\linewidth]{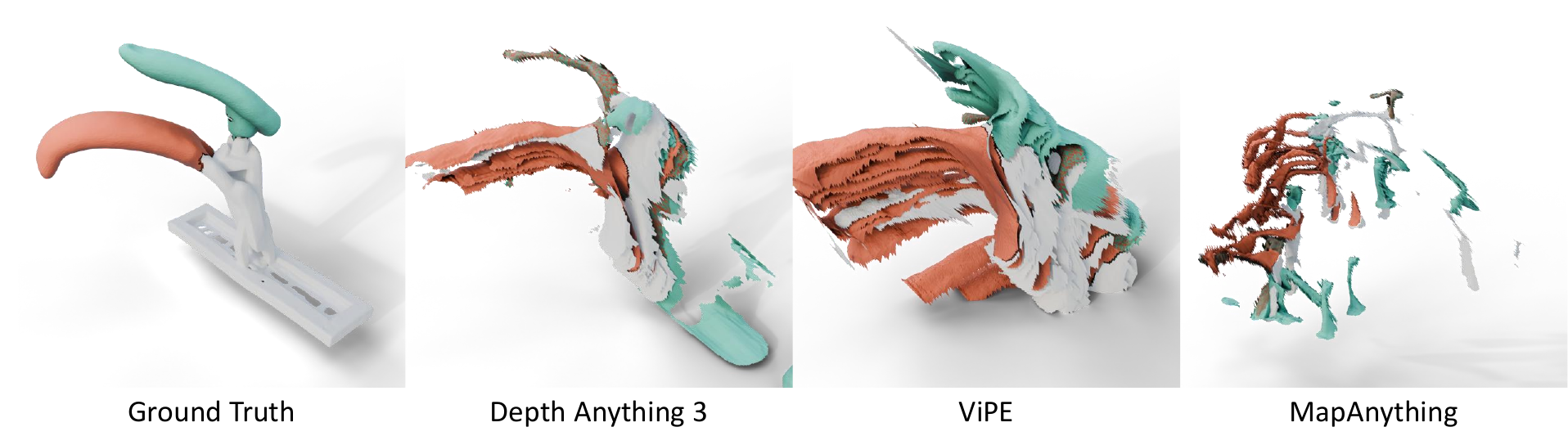}
    \caption{Example results for reconstruction. MapAnything exhibits severe drifting issues as predicted camera poses for different video frames are inaccurate. Other approaches also exhibit significant artifacts. Overall, reconstruction from our egocentric video data is highly challenging for all methods.}
    \label{fig:reconstruction qualitative}
\end{figure}

\mypara{Reconstruction from egocentric videos is misaligned and incomplete.}
\cref{tab:reconstruction} shows that the performance of MapAnything is the worst by a large margin.
This happens due to severe problems with predicted camera intrinsics and extrinsics (see supplement for discussion).
We hypothesize that the large performance gap between MapAnything and Depth Anything 3 (DA3) is due to DA3 being trained on much more data, which covers a more diverse set of scenarios.
DA3 outperforms ViPE slightly and is therefore chosen as the method providing the inputs for the downstream.
From \cref{fig:reconstruction qualitative}, we identify two challenges of reconstruction.
First, the reconstructed meshes across video frames are misaligned.
Second, the meshes are incomplete even though we aggregate results across all frames.
The fusion module is ineffective at aligning reconstructed meshes across frames.
This is likely due to two reasons: 1) small and severely occluded parts do not provide sufficiently reliable feature matches to estimate transformations; and 2) parts that are highly reflective or have featureless textures make finding reliable matches even more difficult.
As a result, we hypothesize that end-to-end, learned 4D reconstruction methods would better fit the task setting, as they may excel at tracking the points and aligning observations into a single state.
3D generative models are another potential solution to generate complete meshes.

\begin{table}[t]
    \caption{Evaluating reconstruction. We report the median value of the chamfer distance and mean value of camera pose prediction error.
    Depth Anything 3 performs the best out of the methods we benchmark. MapAnything severely underperforms due to the camera predictions errors. }
    \centering
    \setlength{\tabcolsep}{8pt}
    \resizebox{\linewidth}{!}{
    \begin{tabular}{@{} l c c c c c @{}}
        \toprule
          Method & Receptor & Effector & Total & Camera & Camera \\
         & CD (m$^2$) & CD (m$^2$) & CD (m$^2$) & RotErr (rad) & TrErr. (m) \\
        \midrule
          MapAnything & 0.408 & 0.733 & 0.572 & 0.960 & 0.663 \\
          Depth Anything 3 & \textbf{0.036} & \textbf{0.019} & \textbf{0.022} & 0.042 & \textbf{0.050} \\
          ViPE & 0.059 & 0.031 & 0.040 & \textbf{0.040} & 0.051\\
         \bottomrule
    \end{tabular}}
    \label{tab:reconstruction}
\end{table}

\begin{figure}[t]
    \centering
    \includegraphics[width=0.95\linewidth]{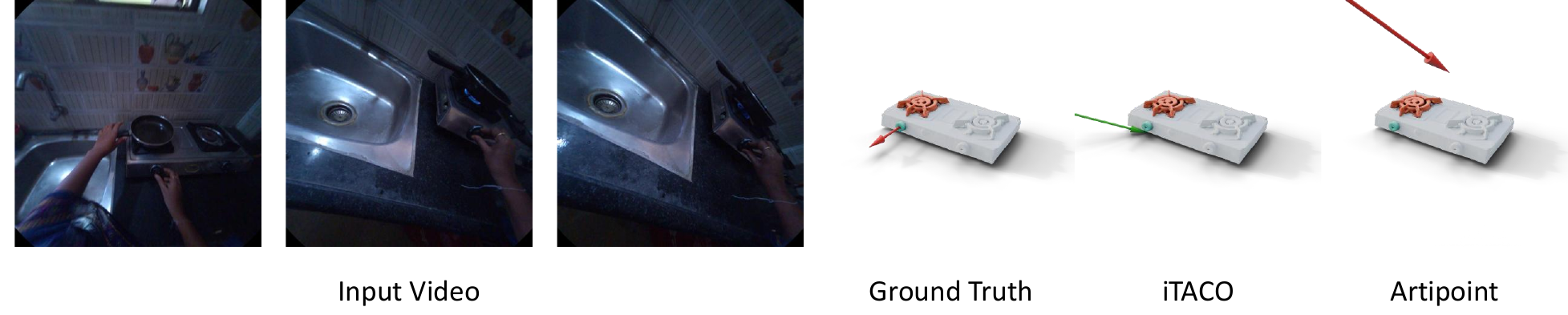}
    \caption{Example results for articulation estimation. In the left example, iTACO predicts incorrect joint types, whereas Artipoint is correct. Both struggle with small parts such as the knob.}
    \label{fig:articulation qualitative}
    \vspace{-12pt}
\end{figure}
\begin{figure*}
    \centering
    \vspace{-12pt}
    \includegraphics[width=0.95\linewidth]{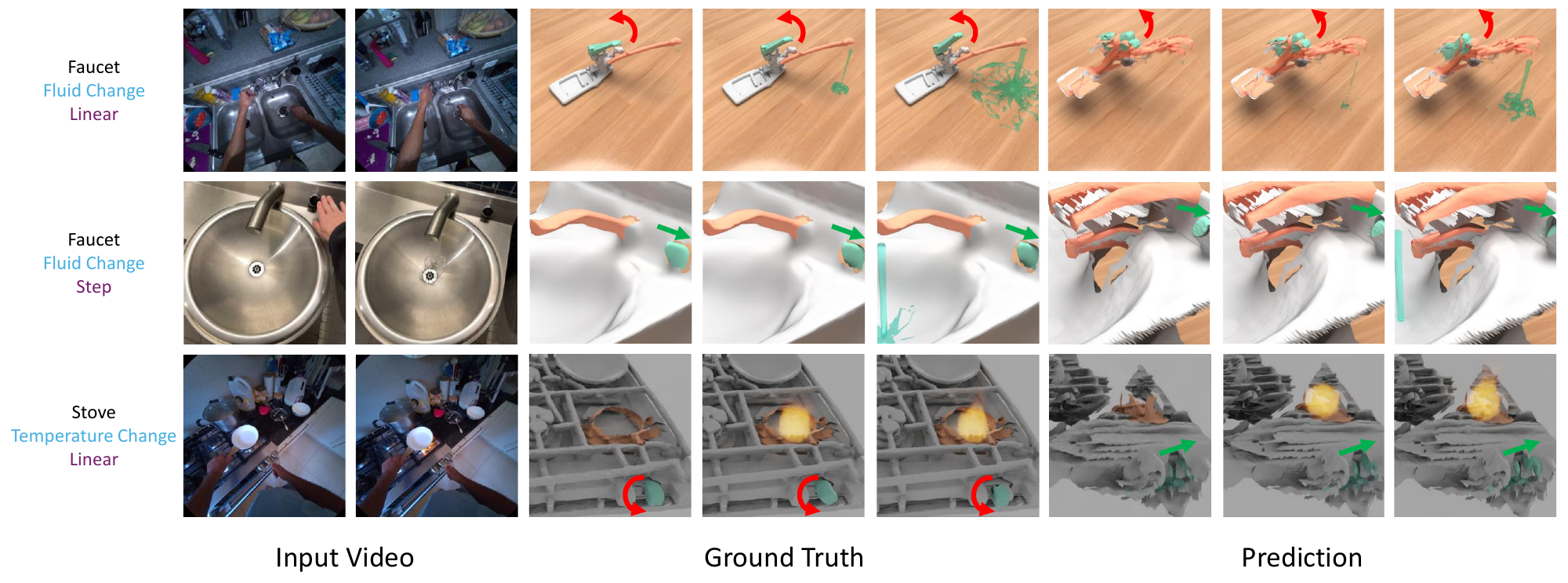}
    \vspace{-1em}
    \caption{Qualitative examples of final outputs. 
    The first two rows show interactive faucets in Genesis~\cite{Genesis}. The last row shows an interactive stove in BEHAVIOR-1K~\cite{li2023behavior}. Colors indicate the receptor (\textcolor{cbteal}{teal}), effector (\textcolor{cborange}{orange}), revolute (\textcolor{red}{red arrow}), and prismatic (\textcolor{green}{green arrow}) joints.}
    \vspace{-12pt}
    \label{fig:final qualitative}
\end{figure*}

\begin{table}[t]
    \caption{Evaluating articulation parameter estimation. We report mean error across videos that successfully go through the pipeline. Artipoint is more accurate than iTACO, but is less robust. Overall, performance for both methods is very low, indicating that articulation estimation is a bottleneck.}
    \centering
    \setlength{\tabcolsep}{8pt}
    \resizebox{\linewidth}{!}{
    \begin{tabular}{@{} l c c c c @{}}
        \toprule
         Method & Axis Err (rad) $\downarrow$ & Origin Err. (m) $\downarrow$ & Type Acc. (\%) $\uparrow$ & Failure (\%) $\downarrow$\\
         \midrule
         
         ArtiPoint & \textbf{1.018} & \textbf{0.331} & \textbf{78.3} & 42.8 \\
         iTACO & 1.024 & 0.758 & 28.8 & \textbf{14.2} \\

         \bottomrule
    \end{tabular}
    }
    \vspace{-12pt}
    \label{tab:articulation}
\end{table}

\mypara{Tracking-based articulation prediction is unreliable.}
From \cref{tab:articulation} and \cref{fig:articulation qualitative} we see that Artipoint generally outperforms iTACO, but suffers from much higher failure rates.
As Artipoint relies on point tracking, it is not surprising that it fails frequently.
Severe occlusions, small parts, and highly dynamic videos, all contribute to poor point tracking.
Consequently, we believe that if a method can estimate articulation parameters from hand motion rather than part motion, it can overcome this challenge, since hands are usually visible in egocentric videos and hand tracking is mature.

\begin{table}[t]
    \caption{Evaluation of function template inference accuracy. We report prediction accuracy for physical effect, mapping, and overall accuracy. A function template is correct if both effect and mapping are correct. We only report accuracy across videos where both receptor and effector segmentation IoUs are larger than 0.5. Among the four different VLMs we benchmarked on this task, Gemini-3-flash performs the best.}
    \centering
    \setlength{\tabcolsep}{8pt}
    \resizebox{\linewidth}{!}{
    \begin{tabular}{@{}l c c c @{}}
        \toprule
         Method & Physical Effect (\%) $\uparrow$ & Mapping (\%) $\uparrow$ & Overall (\%) $\uparrow$ \\
         \midrule

         Gemini 3 Flash & 94.8 & \textbf{96.5} & \textbf{93.1} \\
         GPT-5 mini & 87.9 & 86.2 & 79.3 \\
         Molmo2 8B & 86.2 & 36.2 & 32.7 \\
         Qwen3-VL 8B & \textbf{98.2} & 79.3 & 79.3 \\
          
         \bottomrule
    \end{tabular}
    }
    \vspace{-12pt}
    \label{tab:function}
\end{table}

\mypara{VLMs are good at inferring function templates.}
\Cref{tab:function} shows that Gemini 3 Flash is the most accurate model for this step, GPT-5 mini is the second best, while the two open-source models underperform. Nonetheless, Qwen3-VL achieves somewhat satisfactory performance.
Overall, VLMs are good at inferring function templates.
We believe this is a strong signal towards the suitability of the proposed function template representation for the computational modeling of part functionality.

\subsection{Final Outputs in Simulation}

After predicting the function template, we can compile the template to specific simulators.
Thus, our representation allows creating interactive objects that can be used across simulators.
The supplement discusses the template compilation details.
Example results are in \cref{fig:final qualitative}. 
Overall, the issues described in \cref{subsec:exp_results} are also reflected here.
In the second row, meshes reconstructed from different video frames do not align well due to errors in camera pose estimation and depth estimation.
In the third row, articulation parameter estimation is also challenging, particularly for small parts, resulting in the joint type for the stove knob being misclassified as prismatic by iTACO.
\section{Conclusion}
\label{sec:conclusion}
We proposed \emph{\dataset} a coordinated task, dataset and benchmark for modeling interactive 3D objects from egocentric videos.
To tackle this task, we first presented a function template representation that captures the mapping of cross-part functionality and resulting physical effects.
We showed how this representation enabled portability across simulation platforms through compilation into executable code functions.
We then proposed a system to decompose the overall task into four subtasks: 2D part segmentation, reconstruction, articulation estimation and function prediction.
To evaluate our system comprehensively, we collected and annotated a dataset that provides annotations across egocentric videos and reconstructed 3D objects. 
We benchmarked off-the-shelf components implementing the system on \dataset.
Our results showed that 2D segmentation of small parts, reconstruction from dynamic videos, and reliable articulation estimation are open challenges.
Better and more efficient agentic 2D segmentation frameworks, more accurate end-to-end 4D reconstruction, and more robust articulation estimation based on hand motion are promising directions for future work.

\section*{Acknowledgments}
This work was funded in part by a Canada Research Chair, NSERC Discovery Grant, and enabled by support from the Digital Research Alliance of Canada, and an NVIDIA Academic Grant Award. The authors would like to thank Tianrun Hu from National University of Singapore for collecting data, Jiayi Liu, Xingguang Yan, Austin T. Wang, and Morteza Badali for valuable discussions and proofreading.
{
    \small
    \bibliographystyle{ieeenat_fullname}
    \bibliography{main}

@String(CVPR= {IEEE Conf. Comput. Vis. Pattern Recog.})

@String(ICCV= {Int. Conf. Comput. Vis.})

@String(NIPS= {Adv. Neural Inform. Process. Syst.})

@String(ICLR = {Int. Conf. Learn. Represent.})

@String(AAAI = {AAAI})

@String(CVPR  = {CVPR})

@String(ICCV  = {ICCV})

@String(NIPS  = {NeurIPS})

@String(ICLR  = {ICLR})

@inproceedings{delitzas2024scenefun3d,
  title={{SceneFun3D: Fine-Grained Functionality and Affordance Understanding in 3D Scenes}},
  author={Delitzas, Alexandros and Takmaz, Ayca and Tombari, Federico and Sumner, Robert and Pollefeys, Marc and Engelmann, Francis},
  booktitle=CVPR,
  year={2024}
}

@inproceedings{corsetti2025functionality,
  title={{Functionality understanding and segmentation in 3D scenes}},
  author={Corsetti, Jaime and Giuliari, Francesco and Fasoli, Alice and Boscaini, Davide and Poiesi, Fabio},
  booktitle={Proceedings of the Computer Vision and Pattern Recognition Conference},
  pages={24550--24559},
  year={2025}
}

@article{liu2024grounding,
  title={{Grounding 3D Scene Affordance From Egocentric Interactions}},
  author={Liu, Cuiyu and Zhai, Wei and Yang, Yuhang and Luo, Hongchen and Liang, Sen and Cao, Yang and Zha, Zheng-Jun},
  journal={arXiv preprint arXiv:2409.19650},
  year={2024}
}

@inproceedings{zhang2025open,
  title={{Open-Vocabulary Functional 3D Scene Graphs for Real-World Indoor Spaces}},
  author={Zhang, Chenyangguang and Delitzas, Alexandros and Wang, Fangjinhua and Zhang, Ruida and Ji, Xiangyang and Pollefeys, Marc and Engelmann, Francis},
  booktitle={Proceedings of the Computer Vision and Pattern Recognition Conference},
  pages={19401--19413},
  year={2025}
}

@inproceedings{rotondi2025fungraph,
  title={{FunGraph: Functionality Aware 3D Scene Graphs for Language-Prompted Scene Interaction}},
  author={Rotondi, Dennis and Scaparro, Fabio and Blum, Hermann and Arras, Kai O},
  booktitle={IEEE/RSJ International Conference on Intelligent Robots and Systems},
  year={2025}
}

@inproceedings{todorov2012mujoco,
  title={{MuJoCo: A physics engine for model-based control}},
  author={Todorov, Emanuel and Erez, Tom and Tassa, Yuval},
  booktitle={IEEE/RSJ International Conference on Intelligent Robots and Systems},
  pages={5026--5033},
  year={2012},
  organization={IEEE},
  doi={10.1109/IROS.2012.6386109}
}

@misc{NVIDIA_Isaac_Sim,
author = {{NVIDIA}},
license = {Apache-2.0},
title = {{Isaac Sim}},
url = {https://github.com/isaac-sim/IsaacSim},
version = {5.0.0}
}

@article{Liu2024DWG,
author = {Liu, Weizhou and Li, Jiaze and Chen, Xuhui and Hou, Fei and Xin, Shiqing and Wang, Xingce and Wu, Zhongke and Qian, Chen and He, Ying},
title = {{Diffusing Winding Gradients ({DWG}): A Parallel and Scalable Method for 3D Reconstruction from Unoriented Point Clouds}},
year = {2025},
volume={44},
number={2},
article={19},
numpages = {18},
journal = {ACM Trans. Graph.},
}

@inproceedings{grauman2024ego,
  title={{Ego-Exo4D: Understanding Skilled Human Activity from First- and Third-Person Perspectives}},
  author={Grauman, Kristen and Westbury, Andrew and Torresani, Lorenzo and Kitani, Kris and Malik, Jitendra and Afouras, Triantafyllos and Ashutosh, Kumar and Baiyya, Vijay and Bansal, Siddhant and Boote, Bikram and others},
  booktitle=CVPR,
  year={2024}
}

@misc{carion2025sam3segmentconcepts,
      title={{SAM 3: Segment Anything with Concepts}},
      author={Nicolas Carion and Laura Gustafson and Yuan-Ting Hu and Shoubhik Debnath and Ronghang Hu and Didac Suris and Chaitanya Ryali and Kalyan Vasudev Alwala and Haitham Khedr and Andrew Huang and Jie Lei and Tengyu Ma and Baishan Guo and Arpit Kalla and Markus Marks and Joseph Greer and Meng Wang and Peize Sun and Roman Rädle and Triantafyllos Afouras and Effrosyni Mavroudi and Katherine Xu and Tsung-Han Wu and Yu Zhou and Liliane Momeni and Rishi Hazra and Shuangrui Ding and Sagar Vaze and Francois Porcher and Feng Li and Siyuan Li and Aishwarya Kamath and Ho Kei Cheng and Piotr Dollár and Nikhila Ravi and Kate Saenko and Pengchuan Zhang and Christoph Feichtenhofer},
      year={2025},
      eprint={2511.16719},
      archivePrefix={arXiv},
      primaryClass={cs.CV}
}

@article{ravi2024sam2,
  title={{SAM 2: Segment Anything in Images and Videos}},
  author={Ravi, Nikhila and Gabeur, Valentin and Hu, Yuan-Ting and Hu, Ronghang and Ryali, Chaitanya and Ma, Tengyu and Khedr, Haitham and R{\"a}dle, Roman and Rolland, Chloe and Gustafson, Laura and Mintun, Eric and Pan, Junting and Alwala, Kalyan Vasudev and Carion, Nicolas and Wu, Chao-Yuan and Girshick, Ross and Doll{\'a}r, Piotr and Feichtenhofer, Christoph},
  journal={arXiv preprint arXiv:2408.00714},
  year={2024}
}

@inproceedings{keetha2025mapanything,
  title={{{MapAnything}: Universal Feed-Forward Metric {3D} Reconstruction}},
  author={Nikhil Keetha and Norman M\"{u}ller and Johannes Sch\"{o}nberger and Lorenzo Porzi and Yuchen Zhang and Tobias Fischer and Arno Knapitsch and Duncan Zauss and Ethan Weber and Nelson Antunes and Jonathon Luiten and Manuel Lopez-Antequera and Samuel Rota Bul\`{o} and Christian Richardt and Deva Ramanan and Sebastian Scherer and Peter Kontschieder},
  booktitle={International Conference on 3D Vision},
  year={2026}
}

@misc{record3d,
  author = {{Record3D}},
  title = {Record3D},
  url = {https://record3d.app/},
  version = {1.10.5},
  year = 2025,
  date = {2025-03-11},
}

@misc{polycam,
  author = {{Polycam}},
  title = {Polycam},
  url = {https://poly.cam/},
  version = {5.0.0},
  year = 2025,
  date = {2025-02-26},
}

@misc{Genesis,
  author = {Genesis Authors},
  title = {{Genesis: A Generative and Universal Physics Engine for Robotics and Beyond}},
  month = {December},
  year = {2024},
  url = {https://github.com/Genesis-Embodied-AI/Genesis}
}

@misc{blender,
   title = {Blender - a 3D modelling and rendering package},
   author = {{Blender Online Community}},
   organization = {Blender Foundation},
   year = {2026},
   url = {http://www.blender.org},
}

@InProceedings{halacheva2024articulate3d,
    author    = {Halacheva, Anna-Maria and Miao, Yang and Zaech, Jan-Nico and Wang, Xi and Van Gool, Luc and Paudel, Danda Pani},
    title     = {{Articulate3D: Holistic Understanding of 3D Scenes as Universal Scene Description}},
    booktitle = {Proceedings of the IEEE/CVF International Conference on Computer Vision (ICCV)},
    year      = {2025},
}

@InProceedings{cao2025physx,
  title={{PhysX-3D: Physical-Grounded 3D Asset Generation}},
  author={Cao, Ziang and Chen, Zhaoxi and Pan, Liang and Liu, Ziwei},
  booktitle=NIPS,
  year={2025}
}

@InProceedings{jin2025artvip,
  title={{ArtVIP: Articulated Digital Assets of Visual Realism, Modular Interaction, and Physical Fidelity for Robot Learning}},
  author={Jin, Zhao and Che, Zhengping and Zhao, Zhen and Wu, Kun and Zhang, Yuheng and Zhao, Yinuo and Liu, Zehui and Zhang, Qiang and Ju, Xiaozhu and Tian, Jing and others},
  booktitle=ICLR,
  year={2026}
}

@InProceedings{perrett2025hdepic,
  author    = {Perrett, Toby and Darkhalil, Ahmad and Sinha, Saptarshi and Emara, Omar and Pollard, Sam and Parida, Kranti and Liu, Kaiting and Gatti, Prajwal and Bansal, Siddhant and Flanagan, Kevin and Chalk, Jacob and Zhu, Zhifan and Guerrier, Rhodri and Abdelazim, Fahd and Zhu, Bin and Moltisanti, Davide and Wray, Michael and Doughty, Hazel and Damen, Dima},
  title     = {{HD-EPIC: A Highly-Detailed Egocentric Video Dataset}},
  booktitle = {Proceedings of the IEEE/CVF Conference on Computer Vision and Pattern Recognition (CVPR)},
  year      = {2025},
  month     = {June}
}

@InProceedings{engelbracht2025hoi,
  title={{Hoi!-A Multimodal Dataset for Force-Grounded, Cross-View Articulated Manipulation}},
  author={Engelbracht, Tim and Zurbr{\"u}gg, Ren{\'e} and Wohlrapp, Matteo and B{\"u}chner, Martin and Valada, Abhinav and Pollefeys, Marc and Blum, Hermann and Bauer, Zuria},
  booktitle=CVPR,
  year={2026}
}

@article{arti25werby,
  title={{Articulated Object Estimation in the Wild}},
  author={Werby, Abdelrhman and Buechner, Martin and Roefer, Adrian and Huang, Chenguang and Burgard, Wolfram and Valada, Abhinav},
  journal={Conference on Robot Learning (CoRL)},
  year={2025}
}

@inproceedings{peng2025itaco,
    booktitle = {International Conference on 3D Vision},
    author = {Weikun Peng and Jun Lv and Cewu Lu and Manolis Savva},
    title = {{iTACO: Interactable Digital Twins of Articulated Objects from Casually Captured RGBD Videos}},
    year = {2026}
}

@inproceedings{li2023behavior,
  title={{BEHAVIOR-1K: A Human-Centered, Embodied AI Benchmark with 1,000 Everyday Activities and Realistic Simulation}},
  author={Li, Chengshu and Zhang, Ruohan and Wong, Josiah and Gokmen, Cem and Srivastava, Sanjana and Mart{\'\i}n-Mart{\'\i}n, Roberto and Wang, Chen and Levine, Gabrael and Lingelbach, Michael and Sun, Jiankai and others},
  booktitle={Conference on Robot Learning},
  pages={80--93},
  year={2023},
  organization={PMLR}
}

@article{depthanything3,
  title={{Depth Anything 3: Recovering the Visual Space from Any Views}},
  author={Haotong Lin and Sili Chen and Jun Hao Liew and Donny Y. Chen and Zhenyu Li and Guang Shi and Jiashi Feng and Bingyi Kang},
  journal={arXiv preprint arXiv:2511.10647},
  year={2025}
}

@inproceedings{huang2025vipe,
    title={{ViPE: Video Pose Engine for 3D Geometric Perception}},
    author={Huang, Jiahui and Zhou, Qunjie and Rabeti, Hesam and Korovko, Aleksandr and Ling, Huan and Ren, Xuanchi and Shen, Tianchang and Gao, Jun and Slepichev, Dmitry and Lin, Chen-Hsuan and Ren, Jiawei and Xie, Kevin and Biswas, Joydeep and Leal-Taixe, Laura and Fidler, Sanja},
    booktitle={NVIDIA Research Whitepapers arXiv:2508.10934},
    year={2025}
}

@inproceedings{edstedt2024roma,
title={{RoMa: Robust Dense Feature Matching}},
author={Edstedt, Johan and Sun, Qiyu and Bökman, Georg and Wadenbäck, Mårten and Felsberg, Michael},
booktitle={IEEE Conference on Computer Vision and Pattern Recognition},
year={2024}
}

@article{clark2026molmo2,
      title={{Molmo2: Open Weights and Data for Vision-Language Models with Video Understanding and Grounding}}, 
      author={Christopher Clark and Jieyu Zhang and Zixian Ma and Jae Sung Park and Mohammadreza Salehi and Rohun Tripathi and Sangho Lee and Zhongzheng Ren and Chris Dongjoo Kim and Yinuo Yang and Vincent Shao and Yue Yang and Weikai Huang and Ziqi Gao and Taira Anderson and Jianrui Zhang and Jitesh Jain and George Stoica and Winson Han and Ali Farhadi and Ranjay Krishna},
      year={2026},
      eprint={2601.10611},
      journal={arXiv},
      primaryClass={cs.CV},
}

@article{liu2025videoartgs,
  title={{VideoArtGS: Building Digital Twins of Articulated Objects from Monocular Video}},
  author={Liu, Yu and Jia, Baoxiong and Lu, Ruijie and Gan, Chuyue and Chen, Huayu and Ni, Junfeng and Zhu, Song-Chun and Huang, Siyuan},
  journal={arXiv preprint arXiv:2509.17647},
  year={2025}
}

@article{Qwen3-VL,
      title={Qwen3-VL Technical Report}, 
      author={Shuai Bai and Yuxuan Cai and Ruizhe Chen and Keqin Chen and Xionghui Chen and Zesen Cheng and Lianghao Deng and Wei Ding and Chang Gao and Chunjiang Ge and Wenbin Ge and Zhifang Guo and Qidong Huang and Jie Huang and Fei Huang and Binyuan Hui and Shutong Jiang and Zhaohai Li and Mingsheng Li and Mei Li and Kaixin Li and Zicheng Lin and Junyang Lin and Xuejing Liu and Jiawei Liu and Chenglong Liu and Yang Liu and Dayiheng Liu and Shixuan Liu and Dunjie Lu and Ruilin Luo and Chenxu Lv and Rui Men and Lingchen Meng and Xuancheng Ren and Xingzhang Ren and Sibo Song and Yuchong Sun and Jun Tang and Jianhong Tu and Jianqiang Wan and Peng Wang and Pengfei Wang and Qiuyue Wang and Yuxuan Wang and Tianbao Xie and Yiheng Xu and Haiyang Xu and Jin Xu and Zhibo Yang and Mingkun Yang and Jianxin Yang and An Yang and Bowen Yu and Fei Zhang and Hang Zhang and Xi Zhang and Bo Zheng and Humen Zhong and Jingren Zhou and Fan Zhou and Jing Zhou and Yuanzhi Zhu and Ke Zhu},
	  journal={arXiv preprint arXiv:2511.21631},
      year={2025}
}

@article{openai2025gpt5,
  title={{Introducing GPT‑5}},
  author={OpenAI},
  journal={https://openai.com/index/introducing-gpt-5/},
  year={2025}
}

@misc{geminiteam2025gemini3,
      title={{A new era of intelligence with Gemini 3}}, 
      author={{Gemini Team}},
      year={2025},
      url={https://blog.google/products-and-platforms/products/gemini/gemini-3/},
}

@article{gu2025artisg,
  title={{ArtiSG: Functional 3D Scene Graph Construction via Human-demonstrated Articulated Objects Manipulation}},
  author={Gu, Qiuyi and Sheng, Yuze and Yu, Jincheng and Tang, Jiahao and Shan, Xiaolong and Shen, Zhaoyang and Yi, Tinghao and Liang, Xiaodan and Chen, Xinlei and Wang, Yu},
  journal={arXiv preprint arXiv:2512.24845},
  year={2025}
}

@inproceedings{siddiqui2026shaperrobustconditional3d,
      title={{ShapeR: Robust Conditional 3D Shape Generation from Casual Captures}}, 
      author={Yawar Siddiqui and Duncan Frost and Samir Aroudj and Armen Avetisyan and Henry Howard-Jenkins and Daniel DeTone and Pierre Moulon and Qirui Wu and Zhengqin Li and Julian Straub and Richard Newcombe and Jakob Engel},
      year={2026},
      booktitle=CVPR
}

@inproceedings{pataki2025mpsfm,
  author    = {Zador Pataki and
               Paul-Edouard Sarlin and
               Johannes L. Sch\"onberger and
               Marc Pollefeys},
  title     = {{MP-SfM: Monocular Surface Priors for Robust Structure-from-Motion}},
  booktitle = CVPR,
  year      = {2025}
}

@inproceedings{jiayi2023paris,
  author    = {Liu, Jiayi and Mahdavi-Amiri, Ali and Savva, Manolis},
  title     = {{{PARIS}: Part-level Reconstruction and Motion Analysis for Articulated Objects}},
  year      = {2023},
  booktitle = {Proceedings of the IEEE International Conference on Computer Vision (ICCV)}
}

@inproceedings{weng2024neural,
    title={{Neural Implicit Representation for Building Digital Twins of Unknown Articulated Objects}}, 
    author={Yijia Weng and Bowen Wen and Jonathan Tremblay and Valts Blukis and Dieter Fox and Leonidas Guibas and Stan Birchfield},
    booktitle={Proceedings of the IEEE/CVF Conference on Computer Vision and Pattern Recognition (CVPR)},
    year={2024}
}

@inproceedings{liu2025building,
  title={{Building Interactable Replicas of Complex Articulated Objects via Gaussian Splatting}},
  author={Liu, Yu and Jia, Baoxiong and Lu, Ruijie and Ni, Junfeng and Zhu, Song-Chun and Huang, Siyuan},
  booktitle={International Conference on Learning Representations (ICLR)},
  year={2025},
}

@inproceedings{mandi2024real2code,
  title={{Real2Code: Reconstruct Articulated Objects via Code Generation}},
  author={Mandi, Zhao and Weng, Yijia and Bauer, Dominik and Song, Shuran},
  year      = {2025},
  booktitle   = {International Conference on Learning Representations (ICLR)}
}

@inproceedings{chen2024urdformer,
  title={{URDFormer: A Pipeline for Constructing Articulated Simulation Environments from Real-World Images}},
  author={Zoey Chen and Aaron Walsman and Marius Memmel and Kaichun Mo and Alex Fang and Karthikeya Vemuri and Alan Wu and Dieter Fox and Abhishek Gupta},
  booktitle   = {Robotics: Science and Systems (RSS)},
  year      = {2024},
}

@inproceedings{jiayi2024singapo,
  author    = {Liu, Jiayi and Iliash, Denys and Chang, Angel X. and Savva, Manolis and Mahdavi-Amiri, Ali},
  title     = {{{SINGAPO}: Single Image Controlled Generation of Articulated Parts in Object}},
  year      = {2025},
  booktitle   = {International Conference on Learning Representations (ICLR)}
}

@inproceedings{le2024articulate,
  title={{Articulate-Anything: Automatic Modeling of Articulated Objects via a Vision-Language Foundation Model}},
  author={Le, Long and Xie, Jason and Liang, William and Wang, Hung-Ju and Yang, Yue and Ma, Yecheng Jason and Vedder, Kyle and Krishna, Arjun and Jayaraman, Dinesh and Eaton, Eric},
  booktitle={International Conference on Learning Representations (ICLR)},
  year={2025},
}

@inproceedings{li2025art,
  title     = {{ART: Articulated Reconstruction Transformer}},
  author    = {Li, Zizhang and Zhang, Cheng and Li, Zhengqin and Howard-Jenkins, Henry and Lv, Zhaoyang and Geng, Chen and Wu, Jiajun and Newcombe, Richard and Engel, Jakob and Dong, Zhao},
  booktitle   = CVPR,
  year      = {2025}
}

@inproceedings{yuan2025larm,
author = {Yuan, Sylvia and Shi, Ruoxi and Wei, Xinyue and Zhang, Xiaoshuai and Su, Hao and Liu, Minghua},
title = {{LARM: A Large Articulated Object Reconstruction Model}},
year = {2025},
booktitle = {Proceedings of the SIGGRAPH Asia 2025 Conference Papers}
}

@inproceedings{chen2025freeart3d,
author = {Chen, Chuhao and Liu, Isabella and Wei, Xinyue and Su, Hao and Liu, Minghua},
title = {{FreeArt3D: Training-Free Articulated Object Generation using 3D Diffusion}},
year = {2025},
booktitle = {Proceedings of the SIGGRAPH Asia 2025 Conference Papers}
}

@inproceedings{chen2025artilatent,
author = {Chen, Honghua and Lan, Yushi and Chen, Yongwei and Pan, Xingang},
title = {{ArtiLatent: Realistic Articulated 3D Object Generation via Structured Latents}},
year = {2025},
booktitle = {Proceedings of the SIGGRAPH Asia 2025 Conference Papers}
}

@article{jiang2025phystwin,
    title={PhysTwin: Physics-Informed Reconstruction and Simulation of Deformable Objects from Videos},
    author={Jiang, Hanxiao and Hsu, Hao-Yu and Zhang, Kaifeng and Yu, Hsin-Ni and Wang, Shenlong and Li, Yunzhu},
    journal={ICCV},
    year={2025}
}

@inproceedings{torne2024rialto,
        author    = {Torne, Marcel 
                    and Simeonov, Anthony 
                    and Li, Zechu 
                    and Chan, April 
                    and Chen, Tao 
                    and Gupta, Abhishek 
                    and Agrawal, Pulkit},
        title     = {{Reconciling Reality Through Simulation: A Real-to-Sim-to-Real Approach for Robust Manipulation}},
        booktitle   = {RSS},
        year      = {2024},
      }

@inproceedings{peng2024tiebot,
      title={{TieBot: Learning to Knot a Tie from Visual Demonstration through a Real-to-Sim-to-Real Approach}},
      author={Weikun Peng and Jun Lv and Yuwei Zeng and Hoanan Chen and Siheng Zhao and Jichen Sun and Cewu Lu and Lin Shao},
      booktitle={8th Annual Conference on Robot Learning},
      year={2024},
     }

@article{dan2025xsim,
    title={{X-Sim: Cross-Embodiment Learning via Real-to-Sim-to-Real}}, 
    author={Prithwish Dan and Kushal Kedia and Angela Chao and Edward Weiyi Duan and Maximus Adrian Pace and Wei-Chiu Ma and Sanjiban Choudhury},
    year={2025},
    journal={CoRL}
}

@inproceedings{ning2025prompting,
  title={{Prompting with the Future: Open-World Model Predictive Control with Interactive Digital Twins}},
  author={Ning, Chuanruo and Fang, Kuan and Ma, Wei-Chiu},
  booktitle={RSS},
  year={2025}
}

@article{Escontrela25arXiv_GaussGym,
  journal = {CoRR},
  author = {Escontrela, Alejandro and Kerr, Justin and Allshire, Arthur and Frey, Jonas and Duan, Rocky and Sferrazza, Carmelo and Abbeel, Pieter},
  title = {{GaussGym: An Open-Source Real-To-Sim Framework for Learning Locomotion from Pixels}},
  publisher = {arXiv},
  copyright = {Creative Commons Attribution 4.0 International},
  year = {2025},
  eprint = {2510.15352},
  archiveprefix = {arXiv},
  primaryclass = {cs.RO},
}

@inproceedings{abou-chakra2024physically,
    title={{Physically Embodied Gaussian Splatting: A Realtime Correctable World Model for Robotics}},
    author={Jad Abou-Chakra and Krishan Rana and Feras Dayoub and Niko Suenderhauf},
    booktitle={8th Annual Conference on Robot Learning},
    year={2024}
}

@inproceedings{zhang2025real,
    title={{Real-to-Sim Robot Policy Evaluation with Gaussian Splatting Simulation of Soft-Body Interactions}},
    author={Zhang, Kaifeng and Sha, Shuo and Jiang, Hanxiao and Loper, Matthew and Song, Hyunjong and Cai, Guangyan and Xu, Zhuo and Hu, Xiaochen and Zheng, Changxi and Li, Yunzhu},
    booktitle={ICRA},
    year={2026}
}

@article{buechner2026momasg,
  title={{Articulated 3D Scene Graphs for Open-World Mobile Manipulation}},
  author={Buechner, Martin and Roefer, Adrian and Engelbracht, Tim
          and Welschehold, Tim and Bauer, Zuria and Blum, Hermann
          and Pollefeys, Marc and Valada, Abhinav},
  journal={arXiv preprint arXiv:2602.16356},
  year={2026}
}

@inproceedings{wang2025xsam,
  title={{X-SAM: From Segment Anything to Any Segmentation}},
  author={Wang, Hao and Qiao, Limeng and Jie, Zequn and Huang, Zhijian and Feng, Chengjian and Zheng, Qingfang and Ma, Lin and Lan, Xiangyuan and Liang, Xiaodan},
  booktitle=AAAI,
  year={2026}
}

@article{yuan2025sa2va,
  title={{Sa2VA: Marrying SAM2 with LLaVA for Dense Grounded Understanding of Images and Videos}},
  author={Yuan, Haobo and Li, Xiangtai and Zhang, Tao and Sun, Yueyi and Huang, Zilong and Xu, Shilin and Ji, Shunping and Tong, Yunhai and Qi, Lu and Feng, Jiashi and Yang, Ming-Hsuan},
  journal={arXiv pre-print},
  year={2025}
}

@inproceedings{liu2026visionreasoner,
    title={{VisionReasoner: Unified Reasoning-Integrated Visual Perception via Reinforcement Learning}},
    author={Yuqi Liu and Tianyuan Qu and Zhisheng Zhong and Bohao PENG and Shu Liu and Bei Yu and Jiaya Jia},
    booktitle=ICLR,
    year={2026}
}

@article{kolve2017ai2,
  title={{AI2-THOR: An interactive 3D environment for visual AI}},
  author={Kolve, Eric and Mottaghi, Roozbeh and Han, Winson and VanderBilt, Eli and Weihs, Luca and Herrasti, Alvaro and Deitke, Matt and Ehsani, Kiana and Gordon, Daniel and Zhu, Yuke and others},
  journal={arXiv preprint arXiv:1712.05474},
  year={2017}
}

@inproceedings{wang2019shape2motion,
  title={{Shape2Motion: Joint Analysis of Motion Parts and Attributes from 3D Shapes}},
  author={Wang, Xiaogang and Zhou, Bin and Shi, Yahao and Chen, Xiaowu and Zhao, Qinping and Xu, Kai},
  booktitle={Proceedings of the IEEE/CVF Conference on Computer Vision and Pattern Recognition},
  pages={8876--8884},
  year={2019}
}

@inproceedings{jiang2022opd,
  title={{OPD: Single-view 3D openable part detection}},
  author={Jiang, Hanxiao and Mao, Yongsen and Savva, Manolis and Chang, Angel X},
  booktitle={European Conference on Computer Vision},
  pages={410--426},
  year={2022},
  organization={Springer}
}

@inproceedings{liu2025survey,
  title={{Survey on Modeling of Human-made Articulated Objects}},
  author={Liu, Jiayi and Savva, Manolis and Mahdavi-Amiri, Ali},
  booktitle={Computer Graphics Forum},
  volume={44},
  pages={e70092},
  year={2025},
  organization={Wiley Online Library}
}

@incollection{sherrington2023integrative,
  title={{The Integrative Action of the Nervous System}},
  author={Sherrington, Charles Scott},
  booktitle={Scientific and Medical Knowledge Production, 1796-1918},
  pages={217--253},
  year={2023},
  publisher={Routledge}
}

@article{felzenszwalb2004efficient,
  title={{Efficient Graph-Based Image Segmentation}},
  author={Felzenszwalb, Pedro F and Huttenlocher, Daniel P},
  journal={International Journal of Computer vision},
  volume={59},
  number={2},
  pages={167--181},
  year={2004},
  publisher={Springer}
}

@inproceedings{hu2026hierarchical,
  title={{Hierarchical and Holistic Open-Vocabulary Functional 3D Scene Graphs for Indoor Spaces}},
  author={Hu, Xinggang and Zhang, Chenyangguang and Delitzas, Alexandros and Zhang, Xiangkui and Pollefeys, Marc and Engelmann, Francis and Ji, Xiangyang},
  booktitle={European Conference on Computer Vision},
  year={2026}
}

@article{li2025geosvr,
  title={{GeoSVR: Taming Sparse Voxels for Geometrically Accurate Surface Reconstruction}},
  author={Li, Jiahe and Zhang, Jiawei and Zhang, Youmin and Bai, Xiao and Zheng, Jin and Yu, Xiaohan and Gu, Lin},
  journal={Advances in Neural Information Processing Systems},
  year={2025}
}

@article{li2026mv,
  title={{MV-SAM3D: Adaptive Multi-View Fusion for Layout-Aware 3D Generation}},
  author={Li, Baicheng and Wu, Dong and Li, Jun and Zhou, Shunkai and Zeng, Zecui and Li, Lusong and Zha, Hongbin},
  journal={arXiv preprint arXiv:2603.11633},
  year={2026}
}

@article{
    xiang2025trellis2,
    title={{Native and Compact Structured Latents for 3D Generation}},
    author={Xiang, Jianfeng and Chen, Xiaoxue and Xu, Sicheng and Wang, Ruicheng and Lv, Zelong and Deng, Yu and Zhu, Hongyuan and Dong, Yue and Zhao, Hao and Yuan, Nicholas Jing and Yang, Jiaolong},
    journal={Tech report},
    year={2025}
}

@article{engel2023project,
      title={Project Aria: A New Tool for Egocentric Multi-Modal AI Research},
      author={Jakob Engel and Kiran Somasundaram and Michael Goesele and Albert Sun and Alexander Gamino and Andrew Turner and Arjang Talattof and Arnie Yuan and Bilal Souti and Brighid Meredith and Cheng Peng and Chris Sweeney and Cole Wilson and Dan Barnes and Daniel DeTone and David Caruso and Derek Valleroy and Dinesh Ginjupalli and Duncan Frost and Edward Miller and Elias Mueggler and Evgeniy Oleinik and Fan Zhang and Guruprasad Somasundaram and Gustavo Solaira and Harry Lanaras and Henry Howard-Jenkins and Huixuan Tang and Hyo Jin Kim and Jaime Rivera and Ji Luo and Jing Dong and Julian Straub and Kevin Bailey and Kevin Eckenhoff and Lingni Ma and Luis Pesqueira and Mark Schwesinger and Maurizio Monge and Nan Yang and Nick Charron and Nikhil Raina and Omkar Parkhi and Peter Borschowa and Pierre Moulon and Prince Gupta and Raul Mur-Artal and Robbie Pennington and Sachin Kulkarni and Sagar Miglani and Santosh Gondi and Saransh Solanki and Sean Diener and Shangyi Cheng and Simon Green and Steve Saarinen and Suvam Patra and Tassos Mourikis and Thomas Whelan and Tripti Singh and Vasileios Balntas and Vijay Baiyya and Wilson Dreewes and Xiaqing Pan and Yang Lou and Yipu Zhao and Yusuf Mansour and Yuyang Zou and Zhaoyang Lv and Zijian Wang and Mingfei Yan and Carl Ren and Renzo De Nardi and Richard Newcombe},
      year={2023},
      eprint={2308.13561},
      archivePrefix={arXiv},
}

@inproceedings{delitzas2026funrec,
  title={{Reconstructing Functional 3D Scenes from Egocentric Interaction Videos}},
  author={Delitzas, Alexandros and Zhang, Chenyangguang and Gavryushin, Alexey and Di Mario, Tommaso and Sun, Boyang and Dabral, Rishabh and Guibas, Leonidas and Theobalt, Christian and Pollefeys, Marc and Engelmann, Francis and Barath, Daniel},
  booktitle={IEEE/CVF Conference on Computer Vision and Pattern Recognition (CVPR)},
  year={2026}
}

@inproceedings{guo2026articulat3d,
  title={{Articulat3D: Reconstructing Articulated Digital Twins From Monocular Videos with Geometric and Motion Constraints}},
  author={Guo, Lijun and Zhao, Haoyu and Zhao, Xingyue and Fu, Rong and Zhuang, Linghao and Huang, Siteng and Li, Zhongyu and Zou, Hua},
  booktitle={European Conference on Computer Vision},
  year={2026}
}

@inproceedings{reimers-2019-sentence-bert,
    title = "Sentence-BERT: Sentence Embeddings using Siamese BERT-Networks",
    author = "Reimers, Nils and Gurevych, Iryna",
    booktitle = "Proceedings of the 2019 Conference on Empirical Methods in Natural Language Processing",
    month = "11",
    year = "2019",
    publisher = "Association for Computational Linguistics",
    url = "https://arxiv.org/abs/1908.10084",
}
}

\newpage
\maketitlesupplementary
\renewcommand\thefigure{S\arabic{figure}}
\setcounter{figure}{0}
\renewcommand\thetable{S\arabic{table}}
\setcounter{table}{0}
\renewcommand\theequation{S\arabic{equation}}
\setcounter{equation}{0}
\renewcommand\thelstlisting{S\arabic{lstlisting}}
\setcounter{lstlisting}{0}
\appendix
In the supplementary material, we discuss more details on the formalization and implementation of the function template in \cref{sec:function template details}, comparison with prior datasets in \cref{sec:dataset mesh comparison}, details about the workflow of constructing 3D meshes of \dataset in \cref{sec:data refinement workflow}, and more experiment results and discussions for our benchmark in \cref{sec:exp_supp}.

\section{Function Template Implementation Details}
\label{sec:function template details}

\subsection{Formalization Details}
\label{subsec:function formalization}

Here, we provide in more detail the formalization of the function template representation in terms of mappings between receptor part states and effector part states.
In total, there are 4 possible combinations of state spaces, making up for 8 possible abstractions of mappings. However, we note that only 4 are commonly present in real-world scenarios.
The below list describes each case.

\begin{enumerate}
    \item $\numexp_{\physeff}: s_{\receptor} \rightarrow s_{\effector}$:
    \begin{enumerate}
        \item $s_{\receptor}$ and $s_{\effector}$ are discrete finite sets. In this case, the sizes of $s_{\receptor}$ and $s_{\effector}$ are usually equal. Thus, the mapping is a simple one-to-one mapping. For example, pressing the light switch turns the light on and off. We find that modeling the simplest case of each part admitting two states to be sufficient to cover our data distribution for discrete-to-discrete case, making the resulting mapping a \textbf{binary} function.
        \item $s_{\receptor}$ is a continuous set and $s_{\effector}$ is a discrete finite set. The most common function exemplar for this case from the real-world distribution is a \textbf{step} function. For example, opening the fridge door up to a certain angle triggers the interior lights to turn on.
        \item $s_{\receptor}$ and $s_{\effector}$ are continuous sets. In this case, there are many different valid functions between $s_{\receptor}$ and $s_{\effector}$. However, the real-world examples target simple control. Hence, in most cases, $s_{\receptor}$ and $s_{\effector}$ change proportionally. We use a \textbf{linear} function to approximate this relationship. For example, controlling light intensity proportionally to the rotation of a knob.
    \end{enumerate}
    \item $\numexp_{\physeff}: s_{\receptor} \times s_{\effector} \rightarrow s_{\effector}$:
    \begin{enumerate}
        \item $s_{\receptor}$ and $s_{\effector}$ are discrete finite sets. In this case, $s_{\receptor}$ and $s_{\effector}$ are finite but not necessarily equinumerous. This case is common in many appliances such as washing machines, microwaves, and ovens which are controlled by a few buttons. Such transition functions can be arbitrarily complex. In this work, we consider one of the simplest forms -- the \textbf{cumulative} function. This function maps change in $s_{\receptor}$ to adding a certain value to $s_{\effector}$. For example, pressing a ``plus'' button on an electric stove increases the temperature by 10 degrees.
    \end{enumerate}
\end{enumerate}

Finally, the function template is defined as a function mapping between receptor and effector, which is decomposed to mapping and physical effect. These function templates can be further converted into Python scripts that are executable in specific simulators, such as Mujoco~\cite{todorov2012mujoco}, Isaac Sim~\cite{NVIDIA_Isaac_Sim}, or Genesis~\cite{Genesis}. 

\subsection{Function Template Code Conversion Details}

We first design a code template for each mapping:
\begin{itemize}
    \item Binary: 
        \begin{lstlisting}[language=Python, caption=Pseudocode for binary function, label={lst:binary pseudo code}]
def binary(receptor, effector):
    if receptor.state:
        effector.state = True # Or a real value
    else:
        effector.state = False # Or a real value
        \end{lstlisting}
    \item Step:
        \begin{lstlisting}[language=Python, caption=Pseudocode for step function, label={lst:step pseudo code}]
def step(receptor, effector):
    if receptor.state > THRESHOLD:
        effector.state = True # Or a real value
    else:
        effector.state = False # Or a real value
        \end{lstlisting}
    \item Linear:
        \begin{lstlisting}[language=Python, caption=Pseudocode for linear function, label={lst:linear pseudo code}]
def linear(receptor, effector):
    effector.state = C * receptor.state
        \end{lstlisting}
    \item Cumulative:
        \begin{lstlisting}[language=Python, caption=Pseudocode for cumulative function, label={lst:cumulative pseudo code}]
def cumulative(receptor, effector):
    if receptor.state:
        effector.state = effector.state + delta
        \end{lstlisting}
\end{itemize}

Then, the receptor and effector states are defined based on the target simulator. In our experiments, we demonstrated instantiation of geometry change in Isaac Sim~\cite{NVIDIA_Isaac_Sim}, illumination change and temperature change in BEHAVIOR~\cite{li2023behavior}, and fluid change in Genesis~\cite{Genesis}. 

To instantiate geometry change in Isaac Sim~\cite{NVIDIA_Isaac_Sim}, $s_\receptor$ and $s_\effector$ are the joint state: \texttt{\seqsplit{joint\_state = scene["Object"].data.joint\_pos}}.
Joint states are changed by setting the target joint values: \texttt{\seqsplit{scene["Object"].set\_joint\_position\_target(torch.Tensor([[effector\_target, receptor\_target]]))}}. We show a code example in \cref{lst:isaacsim example}.

\begin{lstlisting}[language=Python, caption=Code example of a step function for geometry change in Isaac Sim, label={lst:isaacsim example}]
joint_state = scene["Microwave"].data.joint_pos # get receptor state
if joint_state[0][0] > 0.015: # check with THRESHOLD
    scene["Microwave"].set_joint_position_target(torch.Tensor([[np.pi / 2, receiver_target]])) # change effector state
\end{lstlisting}

To instantiate illumination change in BEHAVIOR~\cite{li2023behavior}, $s_\receptor$ is either \texttt{\seqsplit{object\_states.ToggledOn}} for binary state or \texttt{object\_states.Joint} for continuous state. $s_\effector$ is either \texttt{light\_bulb.visible} for binary state or \texttt{\seqsplit{light\_bulb.intensity}} for continuous state. The effector state $s_\effector$ is changed by simply assigning a boolean value or a real number, respectively. A code example is shown in \cref{lst:behavior example1}. To instantiate temperature change in BEHAVIOR, $s_\effector$ is \texttt{object\_states.HeatSourceOrSink} and the computation of the temperature value is further defined in this class. A code example is shown in \cref{lst:behavior example2}. The full original source code for \cref{lst:behavior example2} can be found on \href{https://github.com/StanfordVL/BEHAVIOR-1K/blob/88454bd04f75dc57c00ab1f1a00bcde1ff505950/OmniGibson/omnigibson/object_states/heat_source_or_sink.py#L165}{GitHub}.

\begin{lstlisting}[language=Python, caption=Code example of a binary function for illumination change in BEHAVIOR, label={lst:behavior example1}]
if light.states[object_states.ToggledOn].get_value():
    light_bulb.visible = True
else:
    light_bulb.visible = False
\end{lstlisting}

\begin{lstlisting}[language=Python, caption=Code example of a binary function for temperature change in BEHAVIOR, label={lst:behavior example2}]
class HeatSourceOrSink(AbsoluteObjectState, LinkBasedStateMixin, UpdateStateMixin):
    ...
    def _get_value(self):
        # Check the toggle state.
        if self.requires_toggled_on and not self.obj.states[ToggledOn].get_value():
            return False
    
        return True
    ...
\end{lstlisting}

To instantiate fluid change in Genesis~\cite{Genesis}, the receptor state $s_\receptor$ is set to the joint value: \texttt{\seqsplit{joint\_state = faucet.get\_dofs\_position(receptor\_idx)}}. The effector state $s_\effector$ is set to \texttt{droplet\_size} of the emitter, which is a real number. An example code snippet is shown in \cref{lst:genesis example}.

\begin{figure}[t]
\centering
\begin{minipage}{\linewidth}
\begin{lstlisting}[language=Python, caption=Code example of a linear function for fluid change in Genesis, label={lst:genesis example}]
faucet_position = faucet.get_dofs_position(dofs_idx)[0]
droplet_size = change_rate * (faucet_position - joint_limits[dofs_idx[0], 0]) + MIN_DROPLET_SIZE
emitter.emit(
    pos=emitter_position_recentered,
    direction=np.array([0.0, 0.0, -1.0]),
    speed=5,
    droplet_shape="circle",
    droplet_size=droplet_size,
)
\end{lstlisting}
\end{minipage}
\end{figure}

Finally, we use simple strategies to compute a few parameter values needed in the function template. For \texttt{THRESHOLD} in the step mapping in \cref{lst:step pseudo code}, we simply set it to be $\texttt{THRESHOLD}=0.7 \times \max(s_\receptor)$. For \texttt{C} is the linear mapping in \cref{lst:linear pseudo code}, we simply set it to be $\texttt{C}=\frac{\max(s_\effector) - \min(s_\effector)}{\max(s_\receptor) - \min(s_\receptor)}$. For the heat source position in temperature change and the light source position in illumination change, we set them to be the center of the bounding box of the effector mesh. Except for these parameters, we manually specify the \texttt{delta} in \cref{lst:cumulative pseudo code}, minimum and maximum temperature for temperature change, illumination intensity for illumination change, and droplet size for fluid change. We also manually set the emitter position in fluid change.

\section{Qualitative Dataset Comparison}
\label{sec:dataset mesh comparison}
\begin{figure*}[t]
    \centering
    \includegraphics[width=0.95\textwidth]{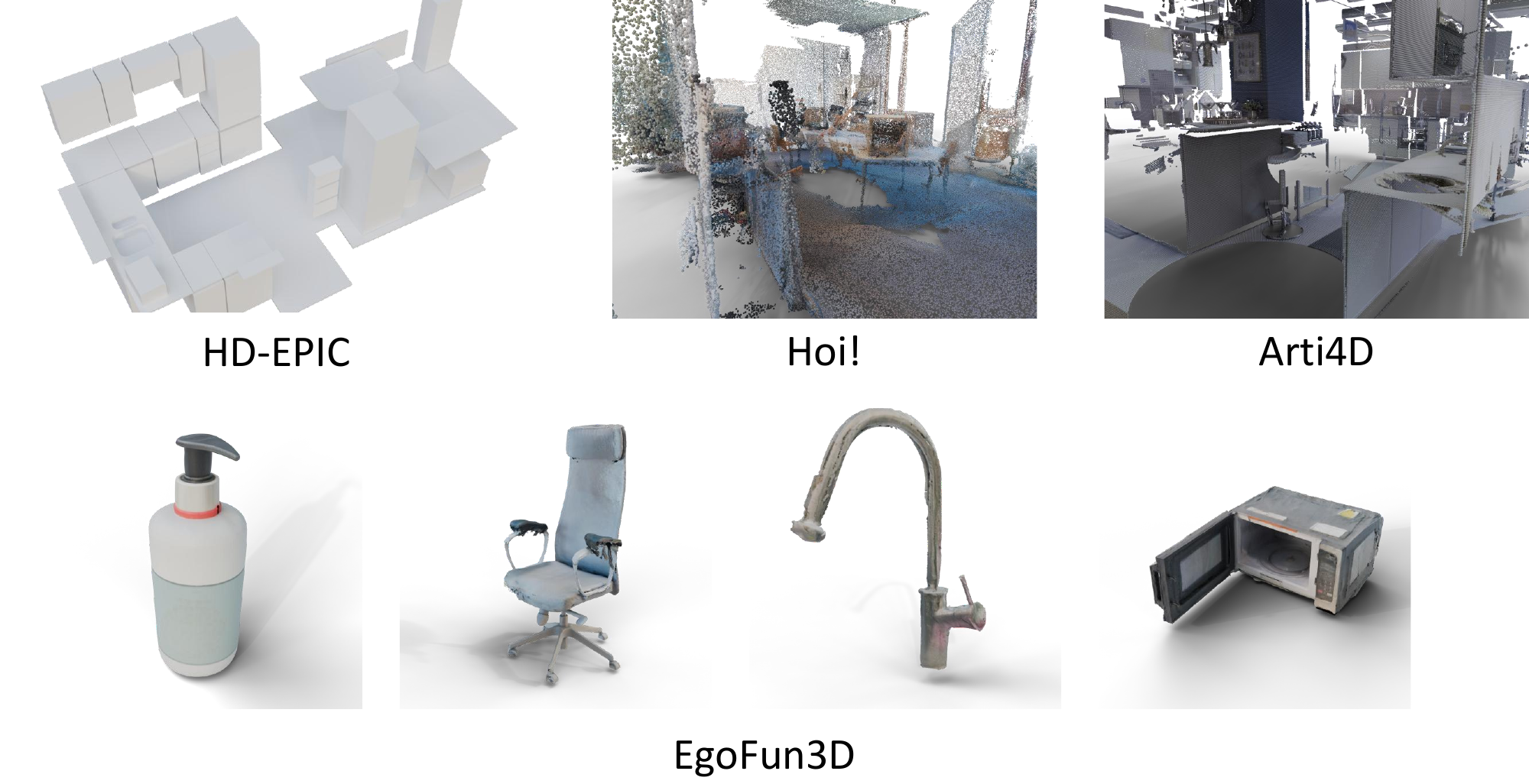}
    \caption{Examples of 3D annotations from our EgoFun3D dataset and prior egocentric human-object interaction video datasets. We see that though some prior datasets provide 3D geometry annotations, their representation is either too coarse (HD-EPIC), or is based on partial and inaccurate point cloud scans (Hoi! and Arti4D). In contrast, EgoFun3D provides complete and accurate ground truth 3D object meshes. }
    \label{fig:dataset mesh comparison}
\end{figure*}
\Cref{fig:dataset mesh comparison} qualitatively illustrates the differences between our EgoFun3D dataset and prior egocentric human-object interaction video datasets that provide 3D annotations, by showing examples of ground truth 3D geometry. HD-EPIC~\cite{perrett2025hdepic} provides only cuboid primitives as 3D geometry and is thus not suitable for evaluating interactive 3D object reconstruction. Hoi!~\cite{engelbracht2025hoi} and Arti4D~\cite{arti25werby} provide scanned point clouds as 3D geometry which are partial and inaccurate, especially for smaller interactive objects like faucets. In contrast, EgoFun3D provides complete and accurate 3D object meshes, which are suitable for evaluating methods that model interactive objects from egocentric video.

\section{Object Geometry Refinement}
\label{sec:data refinement workflow}
\begin{figure*}[t]
    \centering
    \includegraphics[width=\textwidth]{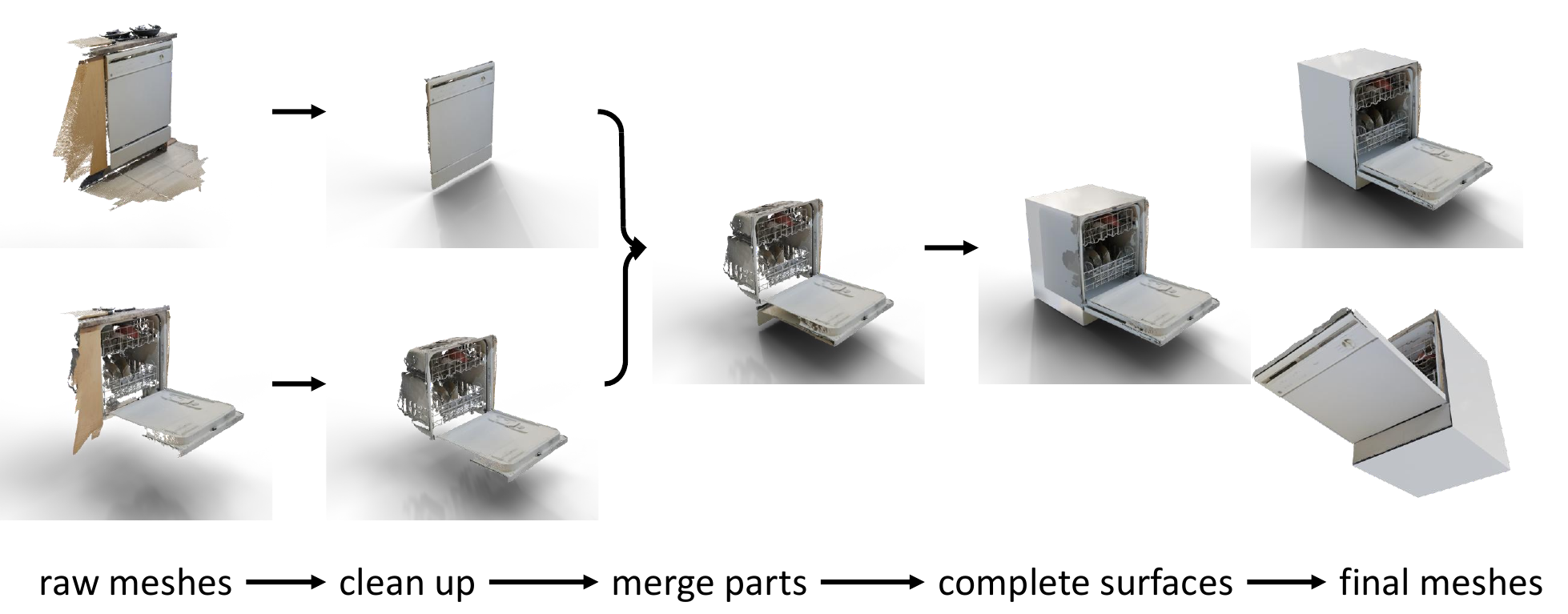}
    \caption{Diagrammatic overview of our 3D ground truth refinement and verification workflow. The multi-stage workflow takes incomplete and inaccurate raw meshes from reconstruction or generation methods, cleans up artifacts, merges different observed states to fully capture interior structure (e.g., dishracks in dishwasher), completes or adds missing object surfaces, and finally verifies dimensions against real-world measurements of the object. The final 3D mesh serves as the ground truth in our EgoFun3D benchmark.}
    \label{fig:mesh refinement workflow}
\end{figure*}
\begin{figure*}[t]
    \centering
    \includegraphics[width=\textwidth]{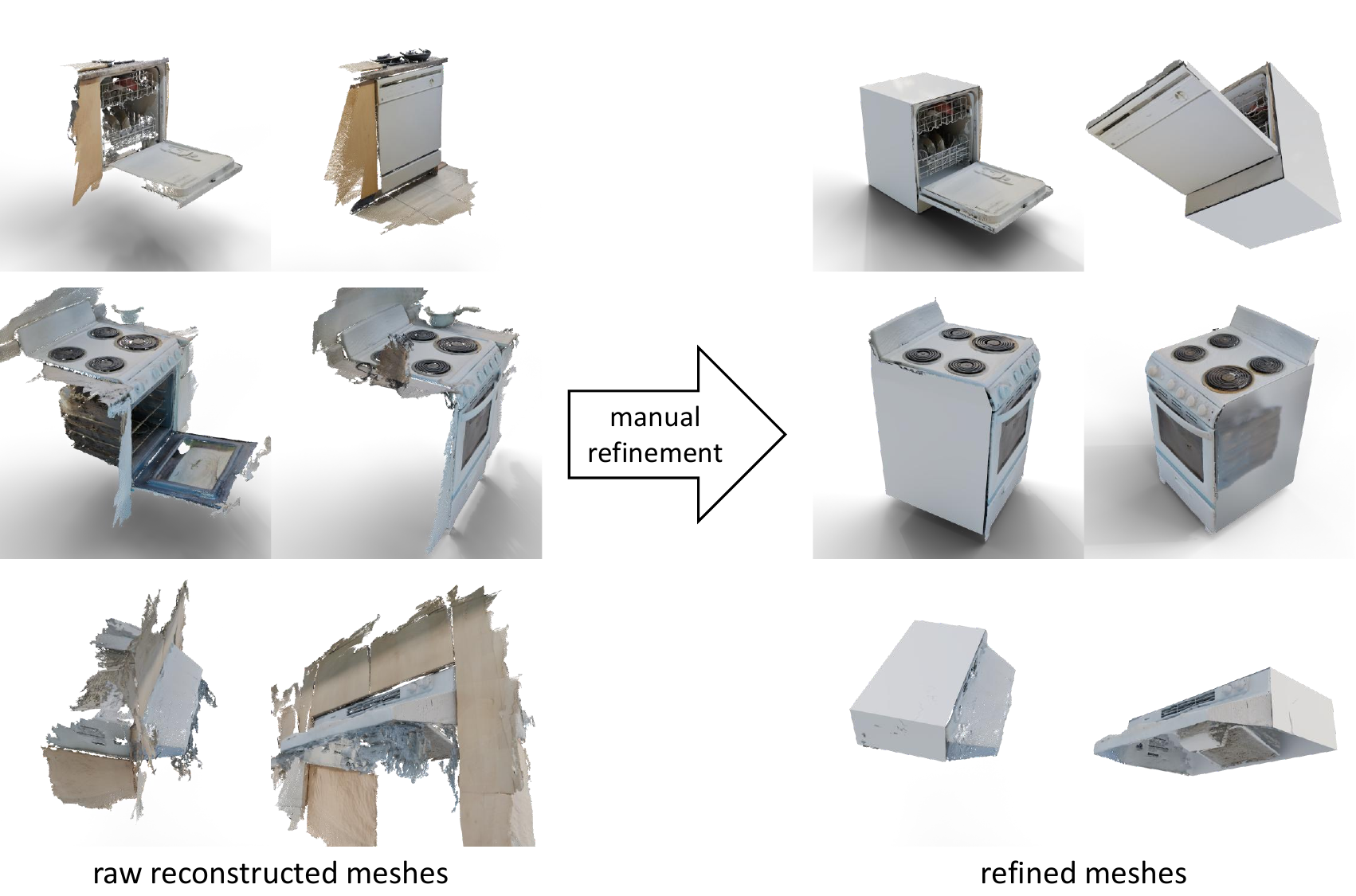}
    \caption{Examples of the input and output for our 3D ground truth refinement and verification workflow. The input meshes on the left and refined and verified to serve as accurate ground truth in our benchmarking. In the second row at the rightmost, the side surface of the range is set to transparent to illustrate the interior structure of the oven.}
    \label{fig:mesh refinement}
\end{figure*}
As described in the main paper, we adopt a multi-stage workflow to create human-verified 3D ground truth geometry for each interactive object in our dataset.

Once we determine the source for the raw 3D mesh according to the workflow from the main paper, we use a Blender-based UI to manually refine and verify the geometry. The following steps are carried out in order: 1) cleaning the raw 3D mesh to remove redundant parts or surfaces due to reconstruction artifacts; 2) merging and registration of parts from different raw mesh sources when necessary, 3) completion or addition of missing surfaces; 4) real-world dimension verification and scaling when necessary. This workflow is illustrated in \cref{fig:mesh refinement workflow}.

Merging is particularly needed when the object has interior features (e.g., dishwasher and range in \cref{fig:mesh refinement}), requiring scanning and reconstruction at the open and closed states of the object to capture interior and exterior structures respectively.

After the missing surface completion and addition stage, the object geometry is compared to real-world measurements of the object and scaled if necessary to conform to the required dimensions. 

\Cref{fig:mesh refinement} shows example qualitative comparisons between the input raw meshes and the refined and verified 3D ground truth meshes after processing with the above workflow. Overall, the final meshes are complete and accurate so they can serve as 3D ground truth in our benchmark.

\section{Additional Experimental Details}
\label{sec:exp_supp}

\subsection{Receptor and Effector Descriptions}

We use Gemini 3 Flash~\cite{geminiteam2025gemini3} for initial video understanding to infer the descriptions of the receptor part and the effector part. We provide the whole video as input, along with a prompt that specifies to identify the part human interacts with and the part that exhibits a reaction to it, as well as a description of spatial relationships of these parts. The exact prompt can be found in \cref{lst:text prompt identification}.

\begin{lstlisting}[
float=t,
floatplacement=htbp,
frame=single,
frameround=tftf,
belowskip=-2\baselineskip,
backgroundcolor=\color{white},
basicstyle=\ttfamily\scriptsize,
columns=fullflexible
keywordstyle=\color{black},
commentstyle=\color{black},
stringstyle=\color{black},
identifierstyle=\color{black},
breakatwhitespace=false,
breaklines=true,
breakindent=0pt,
captionpos=b,
keepspaces=true,
showspaces=false,
showstringspaces=false,
showtabs=false,
caption=Text prompt for receptor and effector identification and description, 
label={lst:text prompt identification},
abovecaptionskip=0pt,
belowcaptionskip=0pt]
"This is a video. Analyze this video and answer the following questions: \
1. Which part of the object receives human action? \
2. Which part of the object reacts to human action? \
Please describe the name and features of the part as well as the spatial relationship with surrounding objects. \
Please only answer in this template: \
{1: {name: xxx, description: aaa}, 2: {name: yyy, description: bbb}} \
Substitue \"xxx\" and \"yyy\" with the name of the part of the object, \"aaa\" and \"bbb\" with the description of the part. \
DO NOT answer any other information."
\end{lstlisting}

\subsection{2D Segmentation}

\mypara{Implementation details.} As the methods we choose rely on a combination of VLM and SAM models, they frequently support a loop where they prompt SAM, evaluate the outputs, refine the prompt and repeat. We allow up to 10 such iterations per frame. We deem this necessary, as the egocentric videos from our dataset are very challenging to segment, while SAM might need multiple rounds of prompt refinement to steer properly. This results in a setup like SAM3 \& Qwen3VL taking a lot of time, as we opt for SAM3 Agent setup from the official SAM3 codebase which involves a particularly long system prompt defining a variety of tool calls a VLM can perform. 

\mypara{Performance in oracle setting.} Additionally, we consider an oracle setting where the ground truth labels for receptor and effector are given. Note that in this setup, longer, spatially-enhanced descriptions of the parts are not available. We find that the trends of relative performance between the selected baselines hold as can be seen in \cref{tab:segmentation_supp}. Ground truth labels lead to generally better performance for the effectors. For the receptors, the performance generally decreases, due to lack of longer descriptions with ground truth labels. Without such spatially-enhanced descriptions, the methods struggle to disambiguate the desired instance in the cases of multiple instances of the same part present in the frame.

\begin{figure*}[t]
    \centering
    \includegraphics[width=0.95\textwidth]{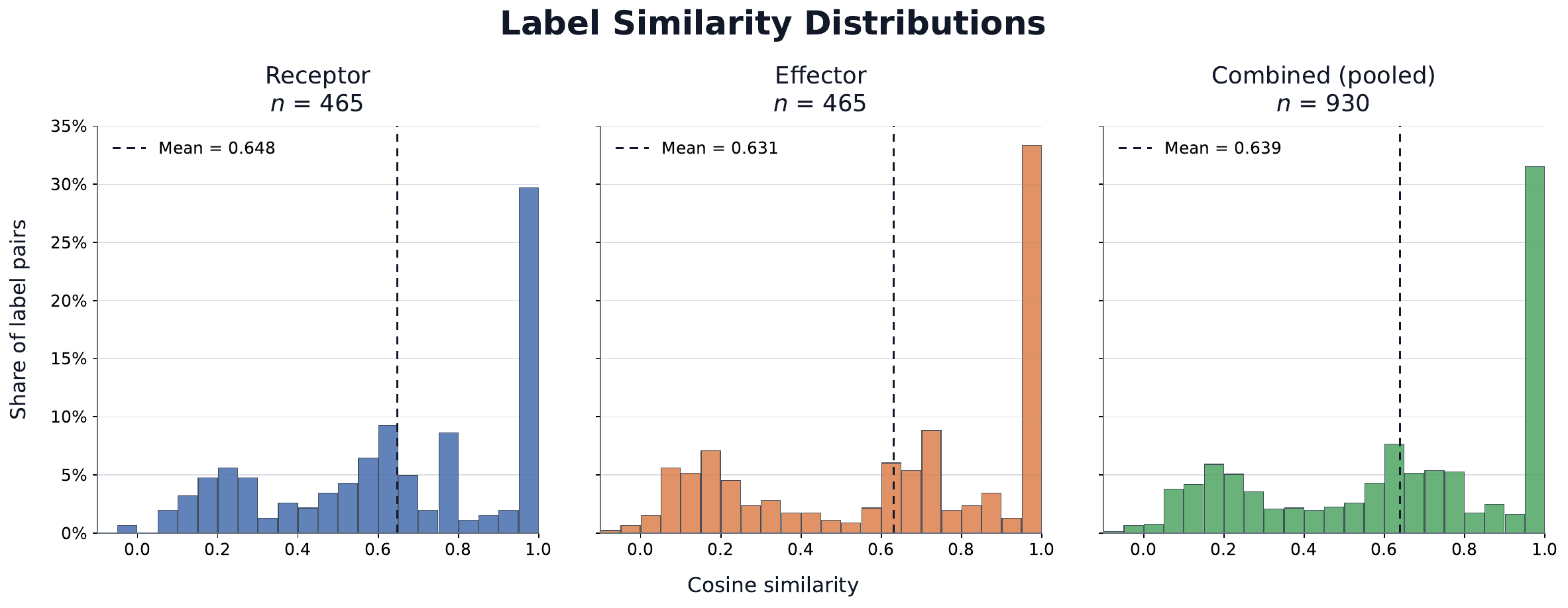}
    \caption{We compute the similarity between the predicted receptor and effector labels and the ground truth labels.}
    \label{fig:similarity distribution}
\end{figure*}

\begin{figure}[t]
    \centering
    \includegraphics[width=0.95\linewidth]{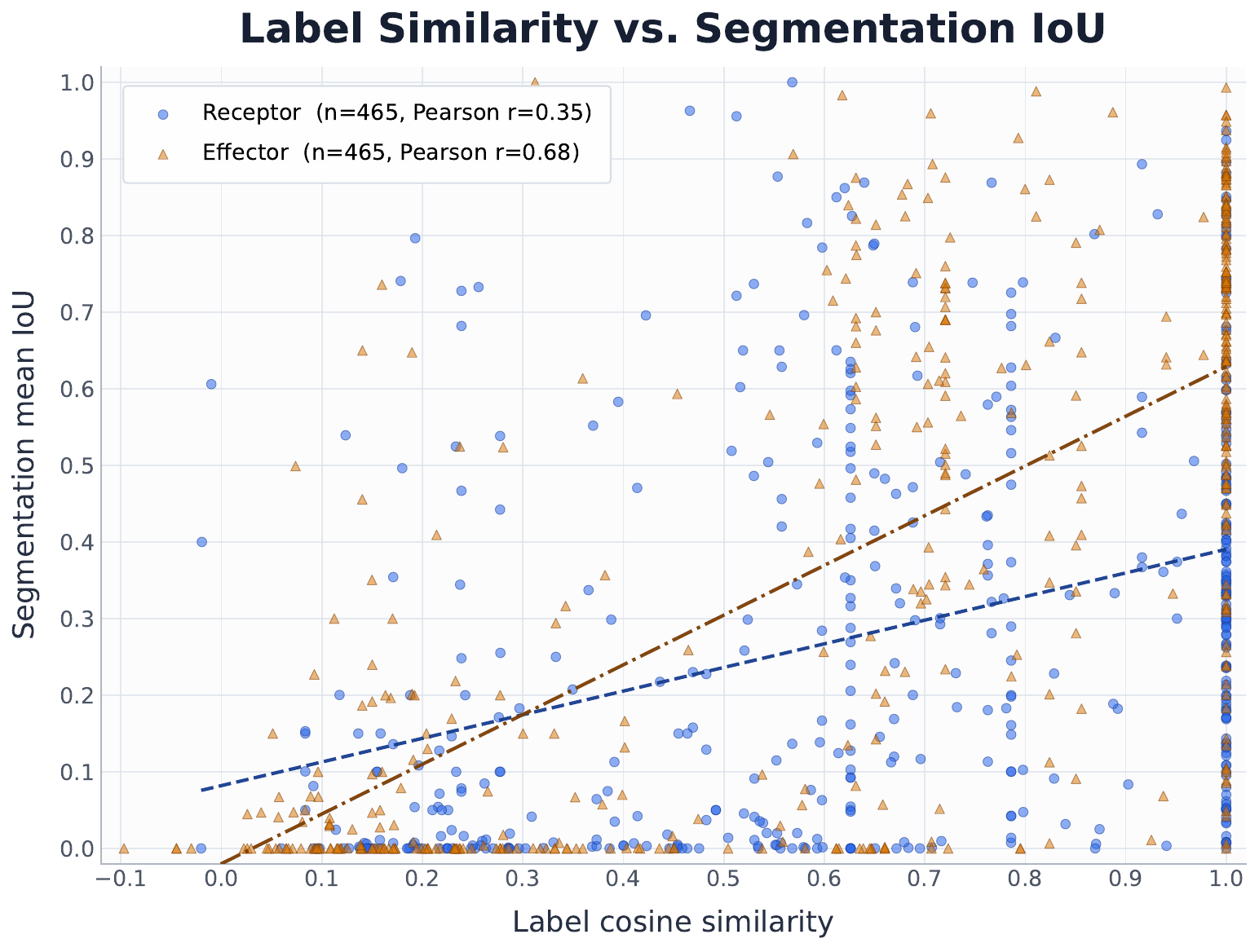}
    \caption{We show the correlation between label prediction similarity and 2D segmentation IoU. We can see that there are positive correlation between the label prediction accuracy and 2D segmentation accuracy.}
    \label{fig:similarity segmentation correlation}
\end{figure}

\mypara{Performance on receptor and effector label prediction.}
We compute the semantic similarity between predicted labels and ground truth labels of receptors and effectors. We compute the embedding features of labels using sentence transformers~\cite{reimers-2019-sentence-bert} and calculate the cosine similarity between embeddings of predicted labels and ground truth labels. The cosine similarity distribution in our dataset is shown in \cref{fig:similarity distribution}. We can see that some labels predicted from Gemini 3 Flash deviate a lot from the correct labels for some videos. We further illustrate that there are positive correlations between the label prediction accuracy and the 2D segmentation accuracy in \cref{fig:similarity segmentation correlation}. Thus, the inaccurate labels predicted by Gemini 3 Flash also contribute to 2D segmentation error.

\begin{table}[t]
    \caption{Evaluation of 2D segmentation performance in the oracle setting, with ground truth receptor and effector labels provided.}
    \centering
    \small
    \setlength{\tabcolsep}{4pt}
    \resizebox{\linewidth}{!}{
    \begin{tabular}{@{} l c c c c c c @{}}
        \toprule
        \multirow{2}{*}{\textbf{Model}}
        & \multicolumn{3}{c}{\textbf{IoU (\%)}}
        & \multicolumn{2}{c}{\textbf{Success (\%)}}
        & \multirow{2}{*}{\textbf{Avg. Run. (s)}} \\

        \cmidrule(lr){2-4} \cmidrule(lr){5-6}

        & Receptor & Effector & Avg.
        & Receptor & Effector
        & \\

        \toprule

        VisionReasoner & 17.3 & 38.2 & 27.7 & 10.1 & 35.5 & 120 \\
        SAM3 \& Qwen3VL & \textbf{34.9} & \textbf{51.8} & \textbf{43.3} & \textbf{27.2} & \textbf{60.0} & 1530 \\
        SAM3 \& Molmo2 & 16.7 & 34.8 & 25.7 & 8.4 & 29.5 & 86 \\
        X-SAM & 3.4 & 11.9 & 7.7 & 0.6 & 6.2 & 25 \\
        SA2VA & 15.1 & 36.0 & 25.6 & 6.0 & 29.5 & 885 \\

        \bottomrule
    \end{tabular}
    }
    \label{tab:segmentation_supp}
\end{table}
\subsection{Reconstruction}

\mypara{Implementation details.} Due to CUDA memory limits, we divide the input video into several chunks when benchmarking MapAnything~\cite{keetha2025mapanything} and Depth Anything 3~\cite{depthanything3}. To stitch reconstruction results for each chunk together, we let the first input video frame of the current video chunk be the same as the last input video frame of the previous video chunk. After reconstructing each chunk, we use the predicted camera pose of the last video frame of the first chunk and the predicted camera pose of the first video frame of the second chunk to compute the transformation. We sequentially transform the reconstruction results of each chunk into the same coordinate system.
In practice, we set the length of each video chunk to be 20 frames.

\begin{figure*}[t]
    \centering
    \includegraphics[width=0.95\textwidth]{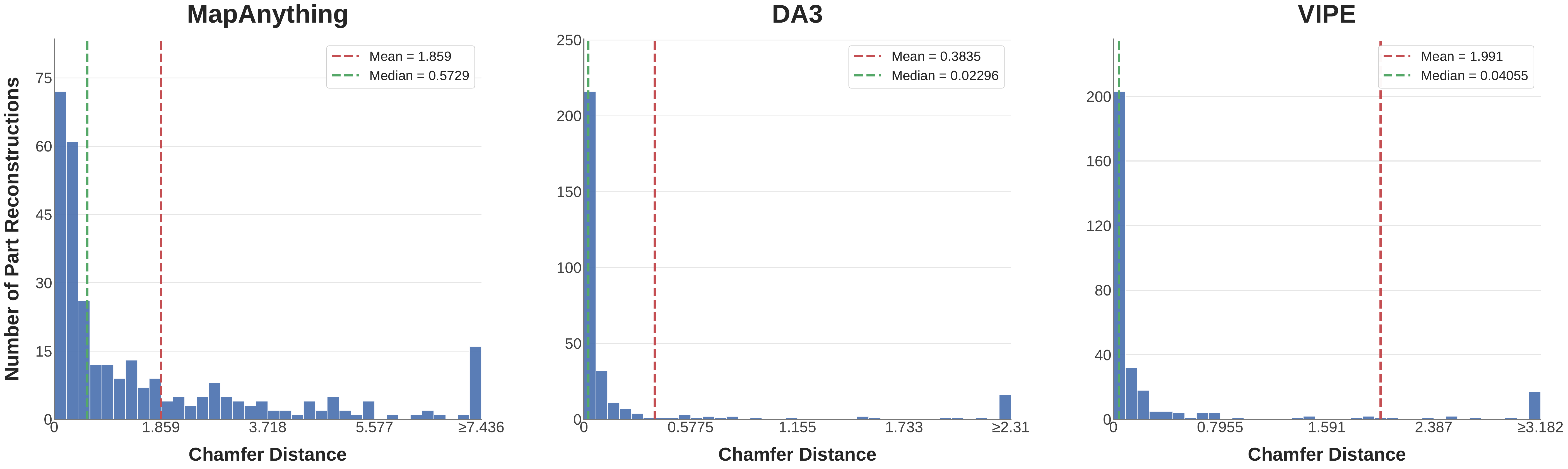}
    \caption{Chamfer Distance distribution on three different methods with RGB video frames and predicted part masks as input. We can see that there are a few extreme values in the reconstruction results. Thus, the median value can better reflect the performance of different reconstruction methods.}
    \label{fig:chamfer dist distribution}
\end{figure*}

\mypara{On the choice of median over mean Chamfer Distance.} We plot the Chamfer Distance distribution on different methods, with video frames and predicted part masks as input in \cref{fig:chamfer dist distribution}. We can see that there are a few extreme values in the reconstruction results, and therefore the mean values are skewed. We also draw the median values and mean values on the charts. We can see that median values lie in the range with most results. Therefore, using the median value can better reflect the overall performance of reconstruction methods.

\begin{table*}
    \caption{Evaluating 4D reconstruction. For inputs including ground truth masks, we report the median value of the chamfer distance across all videos. Among them, Depth Anything 3 performs the best. We also find that the fusion module does not help with reconstruction.
    }
    \centering
    \setlength{\tabcolsep}{8pt}
    \resizebox{\textwidth}{!}{
    \begin{tabular}{@{} l c l l c c c c @{}}
        \toprule
         Input Part Masks & Input Camera & Method & Receptor CD (m$^2$) & Effector CD (m$^2$) & Total CD (m$^2$) & Failure Rate (\%)\\
         \midrule
         \multirow{4}{*}{Ground Truth Masks} & \multirow{4}{*}{\textcolor{green}{\CheckmarkBold}} & MapAnything & 0.051 & 0.026 & 0.035 & 0\\
          & & MapAnything w/o fusion & 0.048 & 0.023 & 0.032 & 0\\
          & & Depth Anything 3 & 0.039 & 0.022 & 0.027 & 4.83\\
          & & Depth Anything 3 w/o fusion & \textbf{0.038} & \textbf{0.020} & \textbf{0.024} & 4.83\\
         \midrule
         \multirow{2}{*}{Predicted Masks} & \multirow{2}{*}{\textcolor{green}{\CheckmarkBold}} & MapAnything & 0.083 & 0.040 & 0.051 & 0 \\
          & & Depth Anything 3 & \textbf{0.062} & \textbf{0.024} & \textbf{0.028} & 2.32 \\
         \midrule
         \multirow{6}{*}{Ground Truth Mask} & \multirow{6}{*}{\textcolor{red}{\XSolidBrush}} & MapAnything & 0.547 & 0.477 & 0.509 & 0\\
          & & MapAnything w/o fusion & 0.454 & 0.361 & 0.421 & 0\\
          & & Depth Anything 3 & 0.020 & 0.014 & 0.016 & 0\\
          & & Depth Anything 3 w/o fusion & \textbf{0.018} & \textbf{0.012} & \textbf{0.014} & 0\\
          & & ViPE & 0.036 & 0.021 & 0.028 & 0\\
          & & ViPE w/o fusion & 0.034 & 0.020 & 0.025 & 0\\
         \midrule
         \multirow{3}{*}{Predicted Mask} & \multirow{3}{*}{\textcolor{red}{\XSolidBrush}} & MapAnything & 0.408 & 0.733 & 0.572 & 0 \\
          & & Depth Anything 3 & \textbf{0.036} & \textbf{0.019} & \textbf{0.022} & 0 \\
          & & ViPE & 0.059 & 0.031 & 0.040 & 0 \\
         \bottomrule
    \end{tabular}}
    \label{tab:reconstruction_supp}
\end{table*}

\begin{figure*}[t]
    \centering
    \includegraphics[width=0.9\textwidth]{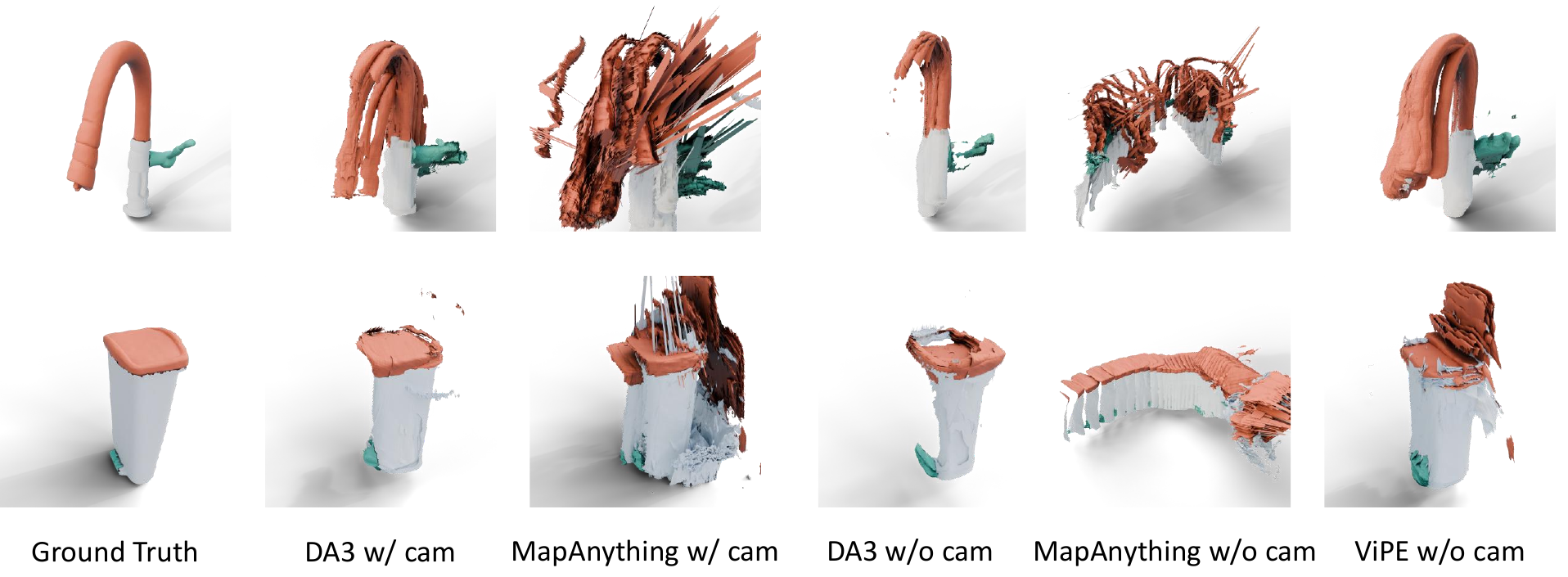}
    \caption{Example results for reconstruction using ground truth part masks. MapAnything exhibits severe drifting issues as predicted camera poses for different video frames are inaccurate.}
    \label{fig:reconstruction qualitative gt mask}
\end{figure*}

\mypara{Ablation study.} We conduct reconstruction experiments across different input modalities, as shown in \cref{tab:reconstruction_supp} and \cref{fig:reconstruction qualitative gt mask}. Comparing experiments with and without ground truth camera parameters input, we find that MapAnything requires accurate camera parameters to achieve high reconstruction accuracy. We also examine the effects of the fusion module. Comparing results between the method with and without the fusion module, we find that the fusion module does not help with reconstruction. We hypothesize that this is due to two reasons: 1) object parts are usually too small to provide reliable feature matching to compute transformations; and 2) many parts are textureless or reflective, increasing the difficulty of finding reliable feature matches.

\subsection{Articulation Estimation}
\begin{table*}[t]
    \caption{Evaluating articulation parameter estimation. We can find that the overall performance for both methods is very inaccurate, indicating that articulation estimation is the bottleneck of this task and the whole system.}
    \centering
    \setlength{\tabcolsep}{8pt}
    \resizebox{\textwidth}{!}{
    \begin{tabular}{@{} l l c c c c @{}}
        \toprule
         Input Part Masks & Method & Joint Axis Err. (rad) $\downarrow$ & Joint Origin Err. (m) $\downarrow$ & Joint Type Acc. (\%) $\uparrow$ & Failure Rate (\%) $\downarrow$\\
         \midrule
         \multirow{2}{*}{Ground Truth Masks} & ArtiPoint & 1.004 & \textbf{0.485} & \textbf{67.058} & 59.036 \\
          & iTACO & \textbf{0.984} & 0.701 & 41.156 & \textbf{29.156} \\
         \midrule
         \multirow{2}{*}{Predicted Masks} & ArtiPoint & \textbf{1.018} & \textbf{0.331} & \textbf{78.333} & 42.851 \\
          & iTACO & 1.024 & 0.758 & 28.888 & \textbf{14.285} \\
         \bottomrule
    \end{tabular}
    }
    \label{tab:articulation_supp}
\end{table*}

We conduct articulation estimation experiments across different input modalities, as shown in \cref{tab:articulation_supp}.
We observe that given ground-truth part masks, performance improves only marginally.

\subsection{Function Template Prediction}
\begin{table*}[t]
    \caption{Evaluating function understanding. We report the accuracy of each aspect and the union of them. Overall predictions are counted as correct when two aspects are predicted correctly. Among the four different VLM, Gemini-3-flash performs the best.
    }
    \centering
    \setlength{\tabcolsep}{8pt}
    \resizebox{\textwidth}{!}{
    \begin{tabular}{@{}l l c c c @{}}
        \toprule
         Input Part Masks & Method & Physical Effect Acc. (\%) $\uparrow$ & Mapping Acc. (\%) $\uparrow$ & Overall Acc. (\%) $\uparrow$ \\
         \midrule
         \multirow{4}{*}{Ground Truth Masks} & Gemini-3-flash & \textbf{90.663} & \textbf{84.520} & \textbf{80.835} \\
          & GPT-5-mini & 84.275 & 76.904 & 69.533 \\
          & Molmo-2-8B & 74.938 & 35.380 & 29.729 \\
          & Qwen3VL-8B-Instruct & 80.835 & 66.830 & 61.916 \\
         \midrule
         \multirow{4}{*}{Predicted Masks} & Gemini-3-flash & 94.827 & \textbf{96.551} & \textbf{93.103} \\
          & GPT-5-mini & 87.931 & 86.206 & 79.310 \\
          & Molmo-2-8B & 86.206 & 36.206 & 32.758 \\
          & Qwen3VL-8B-Instruct & \textbf{98.275} & 79.310 & 79.310 \\
         \bottomrule
    \end{tabular}
    }
    \label{tab:function_supp}
\end{table*}

We evaluate function template prediction with ground truth part masks in \cref{tab:function_supp}.

\subsection{Prompting Details for Function Prediction}

\begin{figure}[t]
    \centering
    \includegraphics[width=0.8\linewidth]{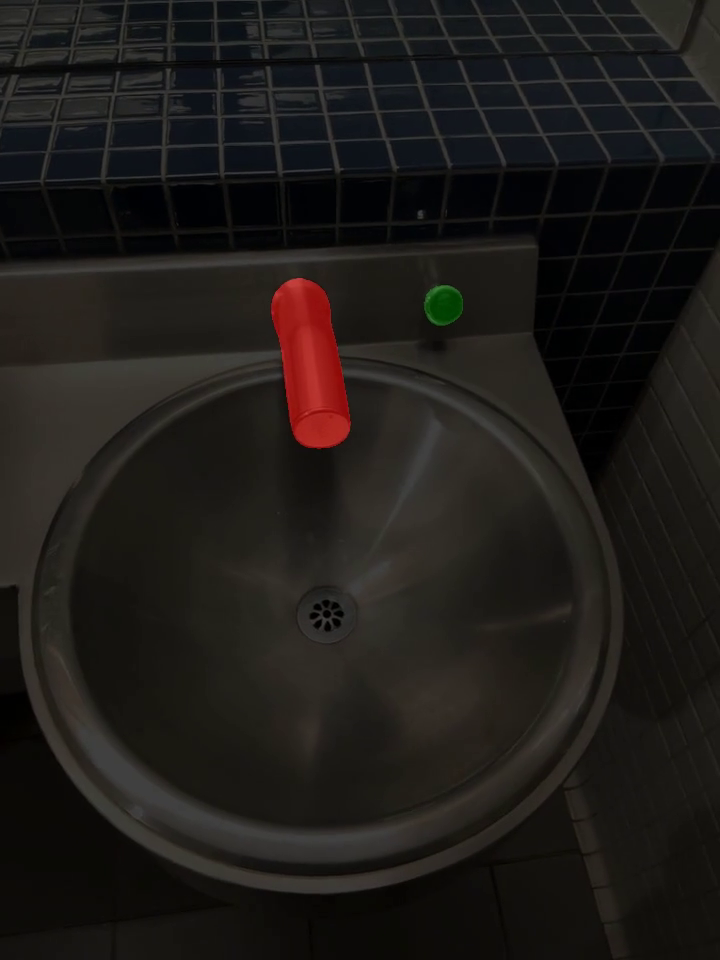}
    \caption{Example video frame we use to prompt VLM for function prediction. The part marked in green is the receptor, and the part marked in red is the effector.
    }
    \label{fig:function prompt video example}
\end{figure}

We prompt VLMs with both text and video for function prediction. The text prompt is shown in \cref{lst:text prompt function}. An example video frame annotated in the way required for function prediction is shown in \cref{fig:function prompt video example}.

\begin{lstlisting}[
float=t,
floatplacement=htbp,
frame=single,
frameround=tftf,
belowskip=-2\baselineskip,
backgroundcolor=\color{white},
basicstyle=\ttfamily\scriptsize,
columns=fullflexible
keywordstyle=\color{black},
commentstyle=\color{black},
stringstyle=\color{black},
identifierstyle=\color{black},
breakatwhitespace=false,
breaklines=true,
breakindent=0pt,
captionpos=b,
keepspaces=true,
showspaces=false,
showstringspaces=false,
showtabs=false,
caption=Text prompt for function prediction, 
label={lst:text prompt function},
abovecaptionskip=0pt,
belowcaptionskip=0pt]
"This is a video. The receptor part marked in green and the effector part marked in red have functional relationship. Function has two attributes: physical effect and numerical relationship. \
Physical effect refers to the physical phenomenon of state change of the effector. We consider four different physical effects: geometry change, illumination change, temperature change, and fluid change. \
If the function results in the change of shape or position of the effector, the physical effect is geometry change. \
If the function results in the change of illumination status of the effector, such as brightness or color, the physical effect is illumination change. \
If the function results in the change of temperature status of the effector, such as becoming hotter or colder, the physical effect is temperature change. \
If the function results in the change of fluid status of the effector, such as fluid flowing in or out, the physical effect is fluid change. \
If the function results in the change of air flow status of the effector, such as air flowing in or out, the physical effect is air flow change. \
Numerical relationship refers to the mathematical relationship between the state of the receptor and the state of the effector. We consider four different numerical function: binary function, step function, linear function, cumulative function. \
If both the states of the receptor and effector have only two possible values, and they are mapped one-to-one, the numerical relationship is binary function. \
For example, the light switch is a binary function, where the state of the receptor (the switch) can be either on or off, and the state of the effector (the light) can also be either on or off. \
If the effector state changes only when the receptor state continuously changes until a certain threshold is reached, the numerical relationship is step function. \
For example, when pressing the button of a microwave to a certain distance, the microwave's door will open. The state of the receptor (the button) continuously changes as you press it, but the state of the effector (the microwave door) only changes when the button is pressed to a certain distance. \
If both states change continuously in a linear manner, the numerical relationship is linear function. \
For example, the volume of water flowing out of a faucet is a linear function of how much you turn the faucet. The state of the receptor (the faucet) and the state of the effector (the water flow) both change continuously, and they have a linear relationship. \
If the effector state accumulates over time as the receptor state changes, the numerical relationship is cumulative function. \
For example, when you press \"increase temperature\" button of a electric induction cooker, the temperature of the cooker will increase a certain value. The state of the receptor (the button) changes as you keep pressing it, and the state of the effector (the cooker temperature) accumulates over time. \
Analyze this video and answer the following questions: \
1. Which physical effect best describes the functional relationship between the red part and green part? Please choose one from the following: \
(a) geometry change \ 
(b) illumination change \
(c) temperature change \
(d) fluid change \
(e) air flow change \
2. Which function best describes the numerical relationship between the state of the red part and the green part? Please choose one from the following: \
(a) binary function \
(b) step function \
(c) linear function \
(d) cumulative function \
Please only answer in this template: \
{\"1\": \"xxx\", \"2\": \"yyy\", \"reason\": \"zzz\"} \
Substitue \"xxx\" and \"yyy\" with the option letter a / b / c / d for each question. Also tell me the reason of choosing each option by substituting \"zzz\". \
DO NOT answer any other information."
\end{lstlisting}

\subsection{Final Results in Simulation}

\begin{figure*}[t]
    \centering
    \includegraphics[width=0.95\textwidth]{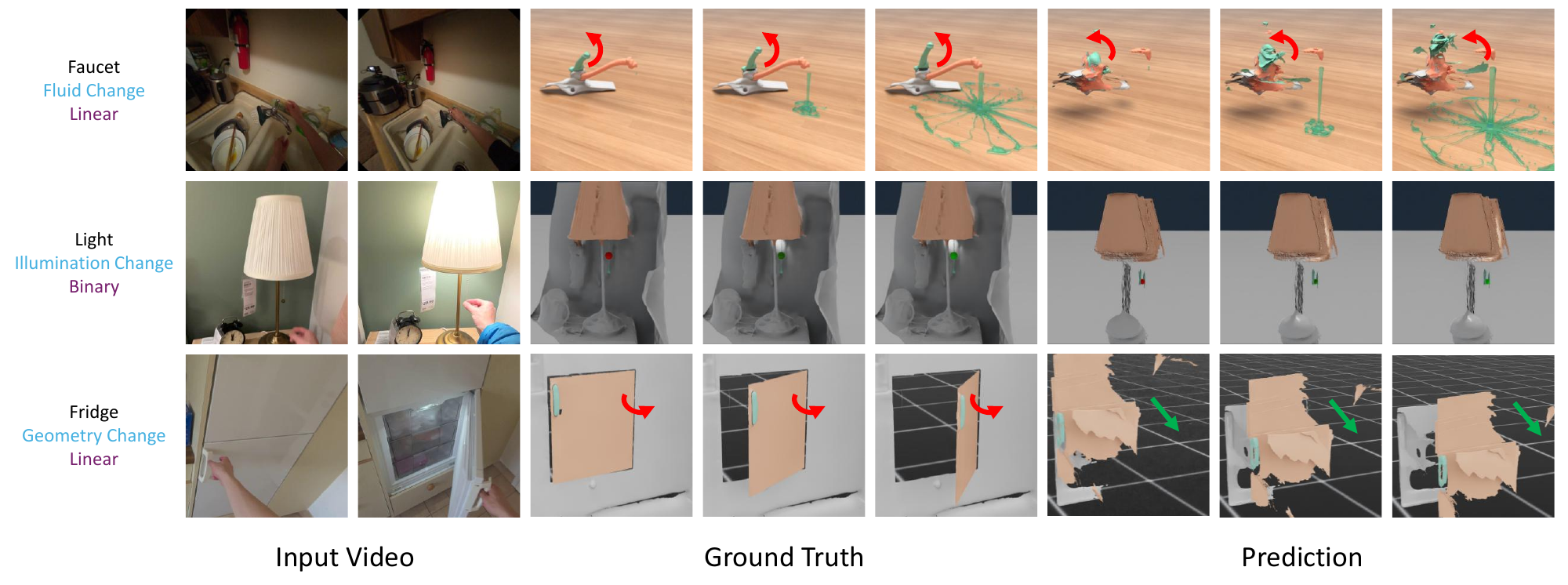}
    \caption{Additional qualitative results of the final outputs of our system. 
    The first row illustrates an interactive faucet in Genesis~\cite{Genesis}. The second row illustrates an interactive lamp in BEHAVIOR-1K~\cite{li2023behavior}. The third row illustrates an interactive fridge door in Isaac Sim~\cite{NVIDIA_Isaac_Sim}. We use teal to indicate receptors and orange to indicate effectors. Red and green arrows represent revolute and prismatic joints respectively.}
    \label{fig:additional final qualitative}
\end{figure*}

We show additional final outputs from out baseline approach in \cref{fig:additional final qualitative}.

\subsection{Additional Discussion of Limitations}

We currently only consider simple functionalities with one receptor and one effector. In reality, there could be multiple receptors and multiple effectors in one part functionality. Also, in \cref{subsec:function formalization}, we do not cover all possible combinations as we did not observe any real part functionalities for some of the cases in our dataset. However, this does not mean such part functionalities do not exist in other real-world scenarios.

Another limitation is that our current function template implementation still requires manual specification of some parameter values, such as the particle emitter positions for fluid physical effects.

\end{document}